\documentclass{article}

     \usepackage[final,nonatbib]{neurips_2020}

\usepackage[utf8]{inputenc} 
\usepackage[T1]{fontenc}    
\usepackage{hyperref}       
\usepackage{url}            
\usepackage{booktabs}       
\usepackage{amsfonts}       
\usepackage{nicefrac}       
\usepackage{microtype}      

\usepackage{amsmath} 
\usepackage{graphicx}
\usepackage{wrapfig}

\renewcommand{\v}[1]{\ensuremath{\mathbf{#1}}} 
 
\newcommand{\abs}[1]{\left| #1 \right|} 
 
\usepackage [english]{babel}
\usepackage [english = american]{csquotes}
\MakeOuterQuote{"}

\title{Deep reconstruction of strange attractors from time series}

\author{%
  William Gilpin\thanks{Code available at: https://github.com/williamgilpin/fnn} \\
  Quantitative Biology Initiative\\
  Harvard University\\
  Cambridge, MA 02138 \\
  \texttt{wgilpin@fas.harvard.edu} \\
}

\begin{document}

\maketitle

\begin{abstract}
Experimental measurements of physical systems often have a limited number of independent channels, causing essential dynamical variables to remain unobserved. However, many popular methods for unsupervised inference of latent dynamics from experimental data implicitly assume that the measurements have higher intrinsic dimensionality than the underlying system---making coordinate identification a dimensionality reduction problem. Here, we study the opposite limit, in which hidden governing coordinates must be inferred from only a low-dimensional time series of measurements. Inspired by classical analysis techniques for partial observations of chaotic attractors, we introduce a general embedding technique for univariate and multivariate time series, consisting of an autoencoder trained with a novel latent-space loss function. We show that our technique reconstructs the strange attractors of synthetic and real-world systems better than existing techniques, and that it creates consistent, predictive representations of even stochastic systems. We conclude by using our technique to discover dynamical attractors in diverse systems such as patient electrocardiograms, household electricity usage, neural spiking, and eruptions of the Old Faithful geyser---demonstrating diverse applications of our technique for exploratory data analysis.
\end{abstract}

\section{Introduction}
\label{intro}

Faced with an unfamiliar experimental system, it is often impossible to know {\it a priori} which quantities to measure in order to gain insight into the system's dynamics. Instead, one typically must rely on whichever measurements are readily observable or technically feasible, resulting in partial measurements that fail to fully describe a system's important properties. These hidden variables seemingly preclude model building, yet history provides many compelling counterexamples of mechanistic insight emerging from simple measurements---from Shaw's inference of the strange attractor driving an irregularly-dripping faucet, to Winfree's discovery of toroidal geometry in the {\it Drosophila} developmental clock \cite{shaw1984dripping,winfree1980}.

Here, we consider this problem in the context of recent advances in unsupervised learning, which have been applied to the broad problem of discovering dynamical models directly from experimental data.  Given high-dimensional observations of an experimental system, various algorithms can be used to extract latent coordinates that are either time-evolved through empirical operators or fit directly to differential equations \cite{champion2019data,sun2019learning,takeishi2017learning,linderman2017bayesian,costa2019adaptive,gilpin2019cellular,pandarinath2018inferring,otto2019linearly}. This process represents an empirical analogue of the traditional model-building approach of physics, in which approximate mean-field or coarse-grained dynamical variables are inferred from first principles, and then used as independent coordinates in a reduced-order model \cite{bar2019learning,gilpin2020learning}. However, many such techniques implicitly assume that the degrees of freedom in the raw data span the system's full dynamics, making dynamical inference a dimensionality reduction problem.

Here, we study the inverse problem: given a single, time-resolved measurement of a complex dynamical system, is it possible to reconstruct the higher-dimensional process driving the dynamics? This process, known as state space reconstruction, is the focus of many classical results in nonlinear dynamics theory, which demonstrate various heuristics for reconstructing effective coordinates given the time history of a system \cite{kennel1992determining,abarbanel1993analysis}. Such techniques have broad application throughout the natural sciences, particularly in areas in which simultaneous multidimensional measurements are difficult to obtain---such as ecology, physiology, and climate science \cite{deyle2011generalized,clark2020nonlinear,ahamed2020capturing,coyle2020ciliate}. However, these embedding techniques are strongly sensitive to hyperparameter choice, system dimensionality, non-stationarity, continuous spectra, and experimental measurement error---therefore requiring extensive tuning and in-sample cross-validation before they can be applied to a new dataset \cite{cobey2016limits,casdagli1991state,deyle2011generalized}. Additionally, current methods cannot consistently infer the underlying dimensionality as the original system, making them prone to redundancy and overfitting \cite{pecora2007unified}. Several of these shortcomings may be addressable by revisiting classical techniques with contemporary methods, thus motivating our study.

Here, we introduce a general method for reconstructing the $d$-dimensional attractor of an unknown dynamical system, given only a univariate measurement time series. We introduce a custom loss function and regularizer, the false-nearest-neighbor loss, that allows recurrent autoencoder networks to successfully reconstruct unseen dynamical variables from time series. We embed a variety of dynamical systems, and we formalize several existing and novel metrics for comparing an inferred attractor to a system's original attractor---and we demonstrate that our method outperforms baseline state space reconstruction methods. We test the consistency of our technique on stochastic dynamical systems, and find that it generates robust embeddings that can effectively forecast the dynamics at long time horizons, in contrast with previous methods. We conclude by performing exploratory analysis of datasets that have previously been hypothesized to occupy strange attractors, and discover underlying attractors in systems spanning earth science, neuroscience, and physiology.

\begin{figure}
  \centering
  \includegraphics[width=.7\linewidth]{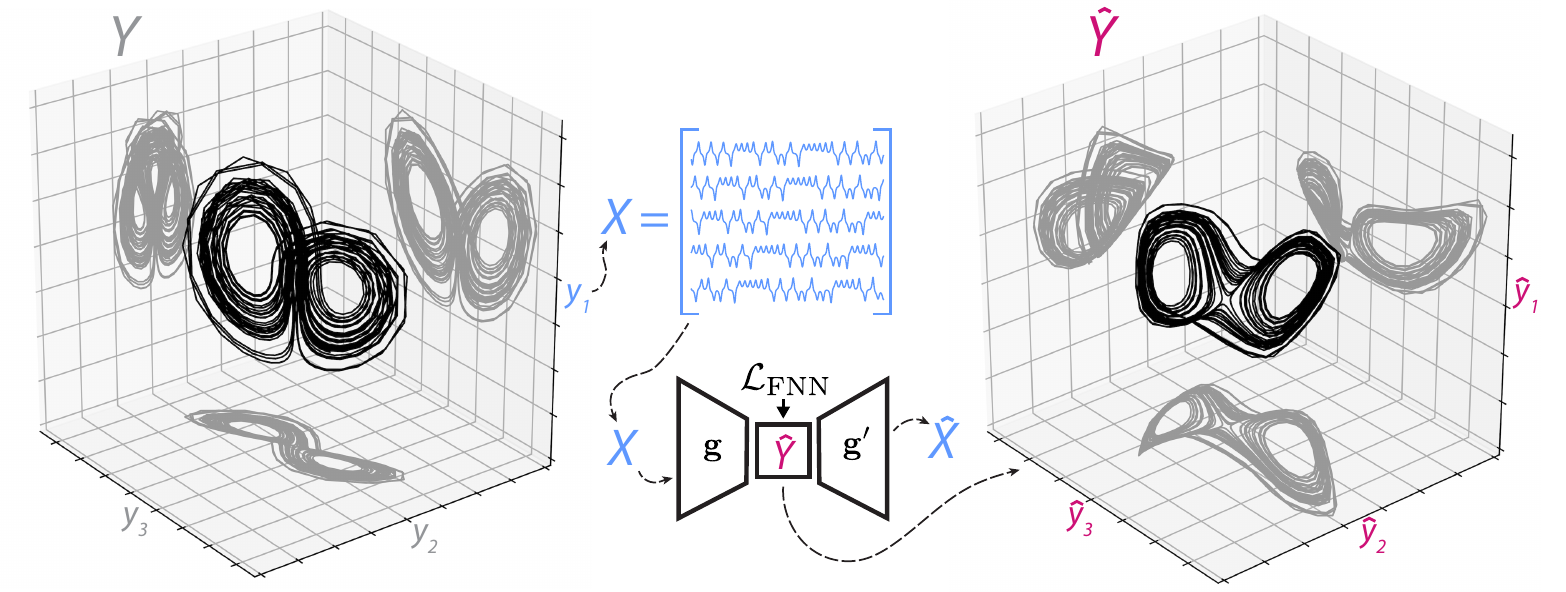}
  \caption{Overview of problem and approach. A univariate time series $y_1(t)$ is observed from a multivariate attractor $Y = [y_1(t)\; y_2(t) \; y_3(t)]$. This signal is converted into a time-lagged Hankel matrix $X$, which is used to train an autoencoder with the false-nearest-neighbor loss $\mathcal{L}_{\text{FNN}}$. The latent variables reconstruct the original coordinates.}
\label{outline}
\end{figure}

\section{Background and Definitions}

Suppose that a $d$-dimensional dynamical system $\dot{\v{y}} = \v{f}(\v y, t)$ occupies an attractor $A$. The time-evolving state variable $\v y$ may be represented abstractly by composition with a flow operator, $\v y(t) = \mathcal F \circ \v y(t_0)$. At any given instant in time, a measurement $\v x(t)$ corresponds to composition with the operator, $\mathcal M$, such that $\v x(t) = \mathcal M \circ \v y(t) = \mathcal M \circ (\mathcal F \circ \v y(t_0))$, where $d_m \equiv \dim{\v x_t}$. We define the data matrix $X = [\v{x}_1^\top\; \v{x}_2^\top\; \cdots\; \v x_N^\top]^\top$ as a collection of $N$ evenly-spaced measurements with timestep $\Delta t$.
Many standard unsupervised embedding techniques for dynamical systems, such as proper orthogonal decomposition or dynamic mode decomposition, implicitly require that $d_m$ is sufficiently large that the measurement operator's basis spans that of the original system, $\text{span}(\mathcal M) \geq \text{span}{(\mathcal F)}$ \cite{takeishi2017learning,mardt2018vampnets}. This condition makes it possible to infer $A$ with sufficient measurements.

Here, we consider the case where high-dimensional time-resolved measurements are unavailable, $\text{span}(\mathcal M) < \text{span}{(\mathcal F)}$, making it more challenging to infer the underlying dynamics. A common example is the univariate case $d_m =1$, such that $X = [x_1\; x_2\; \cdots\; x_N]^\top$. A standard solution in time series analysis is to augment the dimensionality of the measurements via the method of lags, in which the $T$ previous measurements are appended to each timestep to produce a multidimensional surrogate measurement $\v x_i = [x_{i-T}\;\, x_{i - T + 1}\;\, \cdots\;\, x_i]$. In principle, $T$ should be large enough that $x$ (and potentially $\v y$) undergoes sufficient variation to provide information about the dynamics of each component $y_j$ of the underlying system. After augmenting dimensionality with lags, the measurement matrix $X \in \mathbb R^{N\times T}$ has Hankel structure along its diagonals, and here it will serve as the input for an unsupervised learning problem:

We seek a parametric similarity transformation $\hat{\v y} = \v{g} (\v x)$ such that $\hat Y \sim Y$, where $Y \in \mathbb R^{N\times d}$ and $\hat Y \in \mathbb R^{N\times d_E}$. The point set $Y = [\v{y}_1^\top\; \v{y}_2^\top\;  \cdots\; \v y_N^\top]^\top$ corresponds to a finite-duration sample from the true attractor $A$, and the point set $\hat{Y} =  [\hat{\v y}_1^\top\; \hat{\v y}_2^\top\; \cdots\; \hat{\v{y}}_N^\top]^\top$ refers to the embedding of $\v{x}$ at the same timepoints. Because $Y$ (and thus $d$) is unknown, the embedding dimension hyperparameter $d_E$ imposes the number of coordinates in the embedding. For this reason, we seek similarity $\hat Y \sim Y$, rather than $\hat Y = Y$; the stronger condition of congruency $\hat Y \cong \hat Y$ cannot be assured because a univariate measurement series lacks information about the relative symmetry, chirality, or scaling of $Y$. This can be understood by considering the case where the measurement $\v x$ corresponds to a projection of the dynamics onto a single axis, a process that discards information about the relative ordering of the original coordinates.

For general dynamical systems, the embedding function $\v g$ satisfies several properties. For many dynamical systems, the attractor $A$ is a manifold with fractal dimension $d_F \leq d$ (a non-integer for many chaotic systems), which can be estimated from $Y$ using box-counting or related methods. Likewise, $\dim{\hat{\v y} } = d_E$ is usually chosen to be sufficiently large that some embedding coordinates are linearly dependent, and so the intrinsic manifold dimension of the embedded attractor $E$ is less than or equal to $d_E$. Under weak assumptions on $\v f$ and $\mathcal M$, the Whitney embedding theorem states that any such dynamical attractor $A$ can be continuously and invertibly mapped to a $d_E$-dimensional embedding $E$ as long as $d_E > 2 d_F$. This condition ensures that structural properties of the attractor relevant to the dynamics of the system (and thus to prediction and characterization) will be retained in the embedded attractor as long as $d_E$ is sufficiently large.

However, while the Whitney embedding theorem affirms the feasibility of attractor reconstruction, it does not prescribe a specific method for finding $\v g$ from arbitrary time series. In practice, $\v g$ is often constructed using the method of delays, in which the embedded coordinates comprise a finite number of time-lagged coordinates, $\v{g}(\v{x}_i) = [x_{i - d_E \tau}\; \, x_{i - (d_E - 1)\tau}\;\, \cdots\;\, x_i]^\top$. This technique was first applied to experimental data in the context of turbulence and other dissipative dynamical systems \cite{packard1980geometry}, and it is formalized by Takens' theorem, a corollary of the Whitney theorem that states that $\hat Y$ will be diffeomorphic to $Y$ for {\it any} choice of lag time $\tau$ \cite{takens1981detecting}. However, the properties of $\hat Y$ strongly vary with the choice of lag time $\tau$ \cite{abarbanel1993analysis}. Additional theoretical and empirical studies with lagged coordinates suggest that, for many classes of measurement operations $\mathcal M$, it may be possible to construct embeddings $\hat Y$ that are not only diffeomorphic, but also isometric in the sense of preserving local neighborhoods around points on an attractor \cite{sauer1991embedology,durbin2012time,baraniuk2009random,eftekhari2015new,clarkson2008tighter}---a property implicitly required for forecasting and dynamical analysis of reconstructed attractors \cite{deyle2011generalized}. This has led some authors to speculate that certain embedding techniques satisfy the Nash embedding theorem, a strengthening of the Whitney theorem that gives conditions under which an embedding becomes isometric for sufficiently large $d_E$ \cite{yair2017reconstruction,eftekhari2018stabilizing}.

\section{Related Work}

State space reconstruction with lagged coordinates is widely used in fields ranging from ecology, to medicine, to meteorology \cite{kantz2004nonlinear,abarbanel1993analysis,sugihara2012detecting,deyle2011generalized,ahamed2020capturing}. Many contemporary applications use classical methods for determining $\tau$ and $d_E$ for these embeddings \cite{fraser1986independent,kennel1992determining}, although recent advances have helped reduce the method's sensitivity to these hyperparameters \cite{cao1997practical,pecora2007unified}. Other works have explored the use of multiscale time lags inferred via information theoretic \cite{garcia2005multivariate} or topological \cite{tran2019topological} considerations. However, in the presence of noise, time lagged embeddings may generalize poorly to unseen data, thus requiring extensive cross-validation and Bayesian model selection to ensure robustness \cite{cobey2016limits,dhir2017bayesian}.  

Embeddings may also be constructed via singular-value decomposition of the Hankel matrix, producing a set of "eigen-time-delay coordinates" \cite{juang1985eigensystem,broomhead1989time,ghil2002advanced}. These have recently been used to construct high-dimensional linear operators that can evolve the underlying dynamics \cite{brunton2017chaos}. Other methods of constructing $\v g(.)$ include time-delayed independent components \cite{perez2013identification}, Laplacian eigenmaps \cite{erem2016extensions,han2018structured}, Laplacian spectral analysis \cite{giannakis2012nonlinear}, and reservoir computers \cite{lu2018attractor}. Kernel methods for time series also implicitly lift the time series into a higher-dimensional state space \cite{durbin2012time}, either using fixed nonlinear kernels \cite{muller1997predicting,ghahramani1999learning,mirowski2009dynamic}, or trainable kernels comprising small neural networks \cite{wan1993time}. Several recent studies learn embeddings $\v g(.)$ using an autoencoder, an approach we will revisit here \cite{jiang2017state,lusch2018deep,champion2019data,ayed2019learning,ouala2019learning,uribarri2020structure}. Variational autoencoders may be used to model the latent dynamics probabilistically \cite{rangapuram2018deep,karl2016deep,wang2006gaussian}, in which case rank penalties can enforce dynamical sparsity \cite{sun2019learning,she2018reduced,costa2019adaptive}.

Here, we are particularly interested in the related, but distinct, problem of finding time series embeddings that most closely approximate the true dynamical attractor of the underlying system. Accordingly, we seek coordinates that are both predictive and parsimonious, which we quantify with a variety of similarity metrics described below. Our general approach consists of training a stacked autoencoder on the Hankel matrix of the system via a novel, sparsity-promoting latent-space regularizer, which seeks $d_E \approx d$.

\section{Methods}

\subsection{Approach}
\begin{wrapfigure}{r}{0.4\textwidth}
  \centering
    \vspace{-8mm}
 \includegraphics[width=\linewidth]{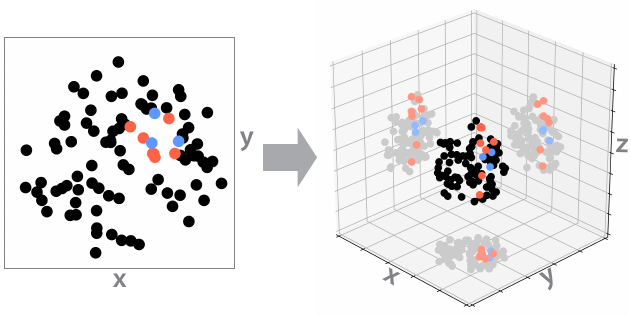}
  \caption{In a 2D projection of a 3D point cloud, false neighbors (red) separate when the system is lifted to 3D.}
  \vspace{-8mm}
  \label{neighbors}
\end{wrapfigure}

We create physically-informative attractors from time series by training an autoencoder, a type of neural network generally used for unsupervised learning \cite{hinton2006reducing}. The encoder portion of the network $\v{g}$ takes as input the Hankel measurement matrix $X$ (or a batch comprising a random subset of its rows), and it acts separately on each row to produce an estimate of the attractor $\hat Y =  \v{g}(X)$, which comprises the latent space of the autoencoder. The decoder $\v{g'}$ takes $\hat Y$ as input, and attempts to reconstruct the input $\hat X = \v{g'}(\hat{Y})$. The encoder and encoder are trained together by composing $\hat X = \v{g'}(\v{g}(X))$, and minimizing the difference between $X$ and $\hat X$. After training the autoencoder, an embedding may be generated either from the training data or unseen test data using only the encoder, $\hat Y =  \v{g}(X)$. To constrain $\v{g'}$ and $\v{g}$, in addition to the reconstruction loss we introduce a novel sparsity-promoting loss $\mathcal{L}_{\text{FNN}}$, which functions as a latent-space activity regularizer. 


We outline our approach in Figure \ref{outline}. In our autoencoder, the hyperparameter $T$ specifies the width of the Hankel matrix (number of time lags) used as input features, while the hyperparameter $L$ (number of latent units) specifies the number of embedding coodinates. However, because of the regularizer, not all $L$ latent units will necessarily remain active after training; our autoencoder thus learns an effective embedding dimension $d_E \leq L$. Rather than culling low-activity latent units to produce an integer $d_E$, we define $d_E$ from our embedding $\hat Y$ post-training as a continuous quantity related to the distribution of relative variance across the $L$ latent units (this is similar to estimating the dimensionality of a PCA embedding from the dropoff in singular values). We aim to have the learned attractor $\hat Y$ match the unseen original attractor $Y$ as closely as possible, including $d_E \approx d$.

Our loss function $\mathcal{L}_{\text{FNN}}$ represents a variational formulation of the false-nearest-neighbors method, a popular heuristic for determining the appropriate embedding dimension $d_E$ when using the method of lags \cite{kennel1992determining}. The intuition behind the technique is that a $d$-dimensional embedding with too few dimensions will have many overlapping points, which undergo large separation when the embedding is lifted to $d+1$. These "false neighbors" only co-localize in $d$ dimensions due to having overlapping projections (Figure \ref{neighbors}). The traditional false-nearest-neighbors technique asserts that the optimal embedding dimension $d_E$ occurs when the fraction of false nearest neighbors first approaches zero as $d$ increases. Here, we modify the technique to apply it iteratively during training: at each optimization step, the false neighbors fraction is estimated from a batch of latent variable activations, and latent variables that fail to substantially decrease the fraction of false neighbors are de-weighted. 

We summarize our regularizer here (see appendix for details): The regularizer $\mathcal{L}_{\text{FNN}}$ accepts as input a batch of $B$ coordinates of dimensionality $L$ corresponding to hidden activations $h \in \mathbb R^{B \times L}$. Here, $h$ corresponds to the network's current estimate of the attractor $\hat Y$ generated from an input comprising $B$ length-$T$ rows randomly sampled from the full measurement matrix $X$. However, we note that our activity regularizer will work for any neural network with hidden layers, independently of the time series embedding problem studied here. Next, pairwise distances in latent space are computed among all points in the batch using only the first $m \leq L$ coordinates, resulting in a dimension-indexed Euclidean distance $D \in \mathbb R^{B \times B \times L}$. This matrix is sorted by column, and the first $K+1$ columns are used to generate a list of $K$ batch indices corresponding to the $K$ nearest neighbors of each point $i$ when only the first $m \in \{1,2,..., L\}$ latent dimensions are considered. Array masking is used to vectorize this computation compared to loop-based implementations in earlier works \cite{kennel1992determining}. The resulting nearest neighbor index matrix is then used to calculate the fraction of $K$-neighbors of each point $i$ that remain the same as the latent index $m$ increases. The fraction of surviving neighbors is then averaged across all points in the batch, producing a batch-averaged fraction of false neighbors $\bar{F} \in \mathbb R^{L}$ that arise at each latent index. The false-neighbors vector $\bar{F} \in \mathbb R^{L}$ then weights an activity regularizer $\mathcal L_{\text{FNN}} = \sum_{m=2}^L (1- \bar{F}_{m}) \bar{h}_m^2$, where $\bar{h}_m$ is the batch-averaged activity of the $m^{th}$ latent unit . Altogether, the loss function for the autoencoder has the form
\[
\mathcal L(X, \hat X, \hat Y) = \| X - \hat X \|^2 + \lambda\, \mathcal{L}_{\text{FNN}}(\hat Y)
\]
where $\|.\|^2$ denotes the Euclidean norm, and $\lambda$ is a hyperparameter controlling the relative strength of the regularizer.

\subsection{Experiments}

{\bf Models.} We illustrate the utility of the loss function $\mathcal{L}_{\text{FNN}}$ across different architectures by using two standard encoder models for all experiments: a single-layer long short-term memory network (LSTM) and a three-layer multilayer perceptron (MLP), each with $L = 10$ latent units. These architectures were chosen because they have a comparable number of parameters ($520$ and $450$, respectively), which is small compared to the minimum of $5000$ timepoints used for all datasets (see supplementary material for additional model details). We obtain comparable results with both models, and we include the MLP results in the appendices. As baseline models, we use eigen-time-delay coordinates (ETD) \cite{broomhead1989time,brunton2017chaos}, time-lagged independent component analysis (tICA) \cite{perez2013identification}, and unregularized replicates of the autoencoders ($\lambda = 0$). In the supplementary material, we also include baseline results for a previously-proposed autoencoder model, comprising a one-layer MLP with $\tanh$ activation \cite{jiang2017state}.

Across all experiments, we only tune the regularizer strength $\lambda$ and the learning rate $\gamma$. Because blind embedding is an unsupervised learning problem, we do not change the network architecture, optimizer, and other hyperparameters. As a general heuristic, we adjust $\lambda$ to be just small enough to avoid dimensionality collapse in the reconstructed attractor (an easily-recognized phenomenon discussed in the next section), and we vary $\gamma$ only to ensure convergence within the constant number of training epochs used for all experiments. For all results, we train five replicate networks with random initializations.

{\bf Datasets.} We study datasets corresponding to several chaotic or quasiperiodic systems: stochastic simulations of the three-dimensional Lorenz "butterfly" attractor, the three-dimensional R\"ossler attractor, a ten-dimensional Lotka-Volterra ecosystem, a three-dimensional quasiperiodic torus, and an experimental dataset corresponding to centroid measurements of a chaotic double pendulum (an effectively four-dimensional system over short timescales) \cite{asseman2018learning}. For all datasets, $5000$ timepoints are used to construct separate Hankel matrices for training and validation of $\v{g}$ and $\v{g'}$, and $5000$ separate timepoints are used as a test dataset for embedding. For each dataset, different replicates or initial conditions are used for train and test partitions whenever possible; for single-series datasets, sets of $5000$ timepoints separated by at least $1000$ timepoints are excerpted to prevent overlap between train, validation, and test. For exploratory analysis of datasets with unknown governing equations, we use datasets corresponding to: temperature measurements of the irregularly-firing "Old Faithful" geyser; a human electrocardiogram; hourly electricity usage measurements for 321 households; and spiking rates for neurons in a mouse thalamus \cite{goldberger2000physiobank,dua2019machine,chaudhuri2019intrinsic}. To ensure consistency across datasets, we downsample all time series to have matching dominant timescales (as measured by leading Fourier mode); otherwise, we apply no smoothing or detrending.

{\bf Evaluation.} Because time series embedding constitutes an unsupervised learning problem, for testing performance against baselines, we train our models by choosing a single coordinate $y_1(t)$ from a known dynamical system $\v y(t)$, which we use to construct a Hankel measurement matrix $X_{\text{train}}$. We then train our autoencoder on $X_{\text{train}}$, and then use it to embed the Hankel matrix of unseen data $X_{\text{test}}$ from the same system, producing the reconstruction $\hat{Y}_{\text{test}}$. We then compare $\hat{Y}_{\text{test}}$ to $Y_{\text{test}}$, a sample of the full attractor at the same timepoints. Because the number of latent coordinates $L$ is the same for all models, but the tested attractors have varying underlying dimensionality $d \leq L$, when comparing $Y$ to $\hat Y$ we lift the dimensionality of $Y$ by appending $L - d$ constant coordinates.

{\bf Metrics.} We use several existing and novel methods to compute the similarity between the original attractor $Y$ and its reconstruction $\hat Y$. We emphasize that this comparison does not occur during training (the autoencoder only sees one coordinate); rather, we use these metrics to assess how well our unsupervised technique reconstructs known systems. We summarize these metrics here (see appendix for additional details):

{\it Pointwise comparison}. Before comparing $\hat Y$ and $Y$, we first apply the Procrustes transform, which applies translation, rotation, and reflection (but not shear) to align $\hat Y$ with $Y$. Because $X$ (and thus $\hat Y$) lacks information about the symmetry and chirality of $Y$, Procrustes alignment prevents relative orientation from affecting subsequent distance calculations. We then calculate the pointwise Euclidean distance, and normalize it to produce the Euclidean similarity. We obtain similar results using the dynamic time warping (DTW) distance, an alternative distance measure for time series \cite{durbin2012time}. Together, these metrics generalize previously-described metrics for comparing strange attractors \cite{diks1996detecting}. 

{\it Forecasting}. We quantify the ability of the reconstructed attractor $\hat{Y}$ to predict future values of the original attractor $Y$, a key property of state space reconstructions used in causal inference \cite{sugihara2012detecting}. We use the cross-mapping forecasting method \cite{sugihara1990nonlinear}; in this algorithm, a simplex comprising the nearest neighbors of each point on $\hat{Y}$ are chosen, and then used to predict future values of each point on $Y$ at $\tau$ timesteps later. We average this forecast across all points, and then scale by the variance, to produce a similarity measure.

{\it Local neighborhoods}. We introduce a novel measure of global neighbor accuracy that describes the average number $\bar\kappa(k)$ of the $k$-nearest neighbors of each point on $\hat Y$ that also fall within the $k$-nearest-neighbors of the corresponding point on $Y$. This quantity is bounded between a perfect reconstruction, $\bar\kappa(k) = k$, and a random sort $\bar\kappa(k) = k^2/N$ (the mean of a hypergeometric distribution). Similar to an ROC-AUC, we compute similarity by summing $\bar\kappa(k)$ from $k=1$ to $k = N -1$, and scale its value between these two limits.

{\it Attractor dimensionality}. A central goal of our approach is determining an appropriate latent dimensionality $d_E$ for the attractor $\hat Y$. We use the variance of each latent coordinate as a continuous measure of its relative activity on the learned attractor, and we compare the variance per index between the embedding $\hat Y$ and full system $Y$. If the full system has $d < L$, we append $L - d$ constant (i.e. zero variance) dimensions to $Y$. We compute the mean square difference between the activity per index of the embedding and of the original system, and we scale this distance to generate a continuous measure of attractor dimension similarity, $\mathcal{S}_\text{dim}$.

{\it Topological features}. We quantify the degree to which $\hat Y$ retains essential structural features of $Y$, such as holes, voids, or the double scrolls of the Lorenz attractor. Following recent work showing that topological data analysis effectively captures global similarity between strange attractors \cite{venkataraman2016persistent,tran2019topological},  we compute the Wasserstein distance between the persistence diagrams of $\hat Y$ and $Y$, which quantifies the presence of different topological features across length scales \cite{edelsbrunner2008persistent}. To produce a similarity measure, we normalize the this distance by the distance between the estimate $\hat Y$ and a null attractor with no salient features.
%

{\it Fractal dimension}. We calculate the similarity between the fractal dimensions of $\hat Y$ and $Y$. We use the correlation fractal dimension---rather than related quantities like Lyapunov exponents or Kolmogorov-Sinai entropy---because it can be calculated deterministically and non-parametrically from finite point sets \cite{grassberger1983measuring}.

\section{Results}

\subsection{Reconstruction of known attractors}

\begin{figure}
\centering
\includegraphics[width=\linewidth]{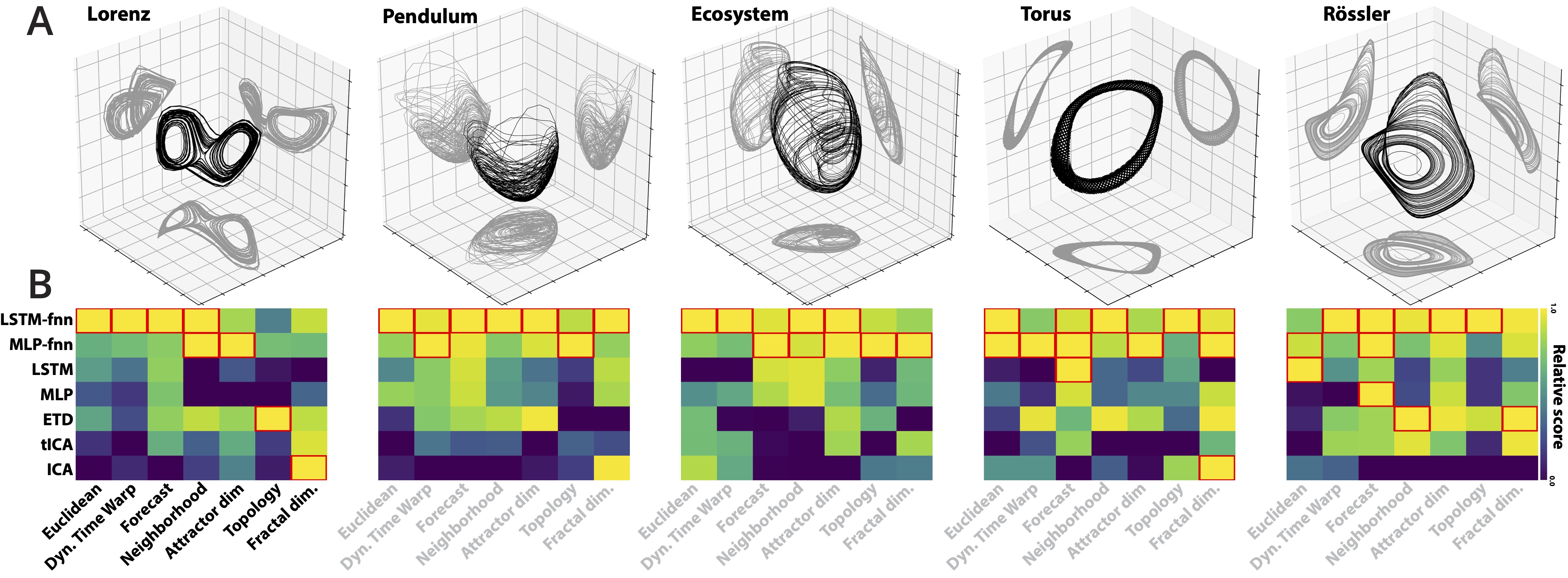}
\caption{
(A) Embeddings produced by the autoencoder with $\mathcal L_{\text{FNN}}$, trained on only the first coordinate of each system. (B) For each system, a variety of baseline embeddings are compared to the original attractor via multiple similarity measures. Hue indicates mean across $5$ replicates scaled by column range, with red boxes indicating column maximum, or values falling within one standard deviation of it. Because distinct similarity metrics have different dynamic ranges, each column has been normalized separately to accentuate differences across models (see appendix for tabular values).
}
\label{attractors}
\end{figure}

Figure \ref{attractors}A shows example embeddings of datasets with known attractors using the LSTM autoencoder with $\mathcal L_{\text{FNN}}$, illustrating the qualitative similarity of the learned embeddings to the true attractors. Figure \ref{attractors}B shows the results of extensive quantitative comparisons between the embeddings and the original attractors, across a variety of measures of attractor similarity (see appendix for raw values).  Compared to baselines, the regularized network either matches or improves the quality of the embedding across a variety of different metrics and datasets. $\mathcal{S}_{\text{dim}}$ consistently improves with the regularizer, demonstrating that $\mathcal L_{\text{FNN}}$ fulfills its primary purpose of generating a latent space with appropriate effective dimensionality $d_E$. Importantly, this effect is not simply due to $\mathcal L_{\text{FNN}}$ indiscriminately compressing the latent space; for the ecosystem dataset, $d = L =10$, and so achieving high $\mathcal{S}_{\text{dim}}$ requires that all latent units remain active after training. We also find that the qualitative appearance of our embeddings, as well as their effective dimensionality $d_E$ depends on $\lambda$ but not $L$ (see appendix). The other metrics encompass measures of cohomology, dynamical similarity, multivariate time series distance, and point cloud similarity, demonstrating that the learned embeddings improve on existing methods in several ways.

Importantly, we observe that the regularized autoencoder nearly always improves on the non-regularized model, suggesting that the regularizer has a clear and beneficial effect on the representations obtained by the model. We hypothesize that the stronger test performance of the regularized model occurs because the regularizer compresses the model more effectively than other latent regularization techniques such as lasso activity regularization (see appendix for comparison) thereby reducing overfitting without sacrificing dynamical information. We emphasize the consistency of our results across these datasets, which span from low-dimensional chaos (Lorenz and R\"ossler attractors), high-dimensional chaos (the ecosystem model), noisy non-stationary experimental data (the double pendulum experiment), and non-chaotic dynamics (the torus).

\subsection{Forecasting noisy time series}

\begin{figure}
\centering
\includegraphics[width=\linewidth]{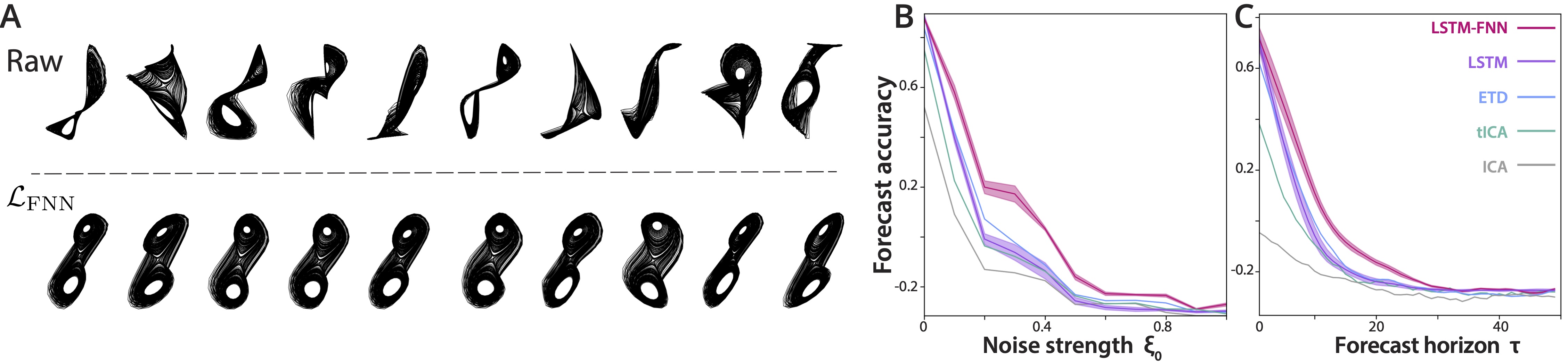}
\caption{
(A) Embeddings of the stochastic Lorenz dataset with and without the false-nearest-neighbors regularizer. Replicates correspond to different random initializations of the Brownian noise force and initial network weights. (B) The cross-mapping forecast accuracy as a function of noise strength $\xi_0$ (with constant $\tau = 20$). (C) The cross-mapping forecast accuracy versus forecasting horizon $\tau$ (with constant $\xi_0 = 0.5$). Standard errors span $5$ replicates.
}
\label{noise}
\end{figure}

Existing attractor reconstruction techniques are often sensitive to noise \cite{cobey2016limits,yap2011stable}. This limitation may be fundamental: Takens' theorem and its corollaries provide no guarantee that a small perturbation to the attractor $Y$ will lead to a small perturbation to $\hat Y$. However, recent theoretical and numerical results have sought attractor reconstruction methods or measurement protocols that remain stable against noise \cite{sauer1991embedology,deyle2011generalized,yap2011stable}. We therefore quantify the robustness of our technique to noise by performing a series of simulations of the Lorenz equations that include time-dependent forcing by uncorrelated Brownian motion. We vary the relative amplitude of the the noise term, and then train separate models for each amplitude. We use the same hyperparameters as for the case without noise, as described above. Figure \ref{noise}C shows the cross-mapping forecasting accuracy as a function of forecasting horizon, $\tau$, and the relative noise amplitude, $\xi_0 \in [0, 1]$ \cite{sugihara1990nonlinear}. Consistent with the results for the attractor similarity measures, we find that the prediction accuracy decays the slowest for the regularized LSTM model, and that the advantage of the regularized model is more pronounced at long forecasting horizons. Moreover, when we train replicate networks with different random initializations (Figure \ref{noise}A), we find that the regularized models consistently converge to similar sets of coordinates---suggesting that our method successfully identifies the salient signal in a noisy time series, and finds a general solution independent of the noise or initial weights.

\subsection{Inferring the dimensionality of an attractor}

\begin{figure}
  \centering
 \includegraphics[width=0.6\linewidth]{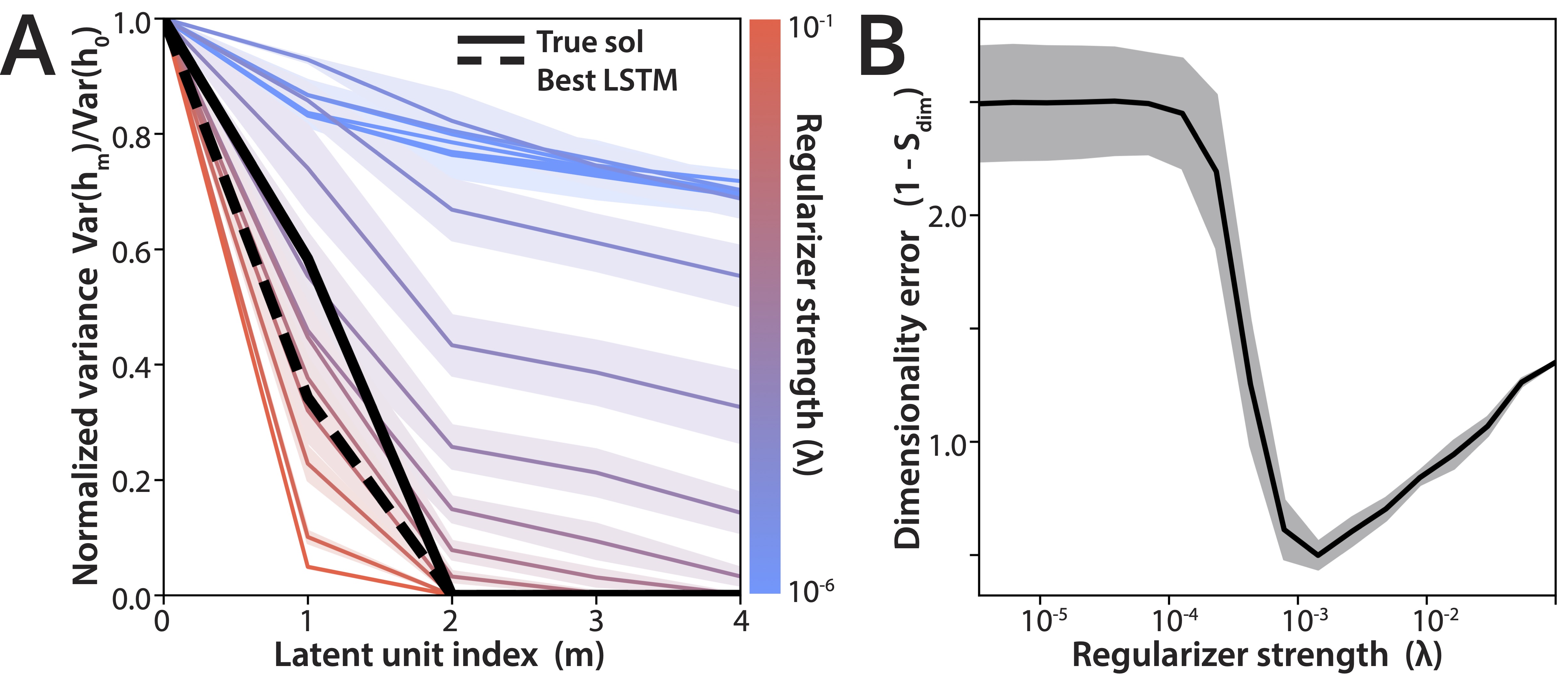}
  \caption{(A) The final distribution of latent variances for replicate networks trained on the Lorenz system, parametrized by regularizer strength; the normalized variance of the final best-performing LSTM (dashed black line) and the variance per coordinate of the full Lorenz system (solid black line) are overlaid. (B) The dimensionality error $1 - \mathcal{S}_{\text{dim}}$ versus $\lambda$. Standard errors span $5$ replicates.}
  \label{reg}
\end{figure}

We next investigate the effect of the regularizer strength $\lambda$ on the embedding. Figure \ref{reg}A shows the effect of increasing the regularizer strength on the variance of the activations of the $L=10$ ranked latent coordinates for embeddings of the Lorenz dataset. Identical experiments with the MLP model are included in the appendix. As $\lambda$ increases, the distribution of activation across latent variables develops increasing right skewness, eventually producing a distribution of activations similar to that of weighted principal components. Figure \ref{reg}B shows the final dimensionality error $1 - \mathcal{S}_{\text{dim}}$ for replicate networks trained with different regularizer strengths.  The plots show that the dimensionality accuracy of the learned representation improves as long as $\lambda$ is greater than a threshold value. However, the error begins to increase if $\lambda$ becomes too large, due to the learned attractor becoming overly flattened, and thus further from the correct dimensionality. This nonlinearity implies a simple heuristic for setting $\lambda$ for an unknown dataset: keep increasing lambda until the effective dimensionality of the latent space rapidly decreases, and then vary it no further.

\subsection{Exploring datasets with unknown attractors}

\begin{figure}
  \centering
 \includegraphics[width=\linewidth]{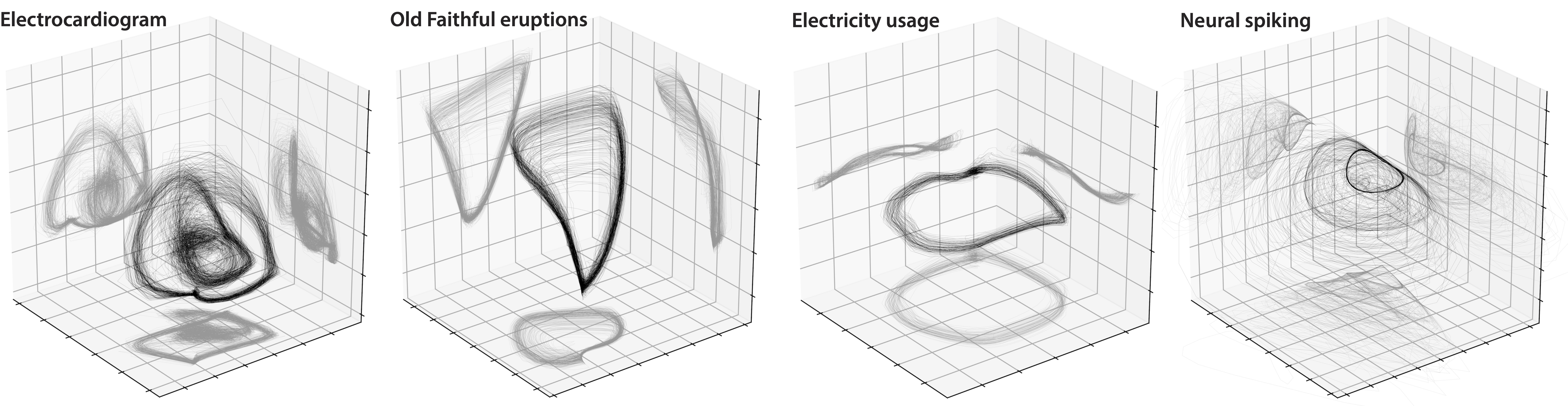}
  \caption{Embeddings of an electrocardiogram ($160$ heartbeats), temperature measurements of the erupting "Old Faithful" geyser in Yellowstone National Park ($200$ eruptions), average electricity usage by $321$ households ($200$ days), and neural spiking in a mouse thalamus.}
  \label{demos}
\end{figure}

To demonstrate the potential utility of our approach for exploratory analysis of unknown time series, we next embed several time series datasets for which the governing equations are unknown, but for which low-dimensional attractors have previously been hypothesized. Figure \ref{demos} shows embeddings of various systems using the FNN loss with the LSTM model. For all systems, a different training dataset is used to construct $\v g(.)$ than the testing dataset plotted. Several qualitative features of the embedded attractors are informative. For the electrocardiogram dataset, the model successfully creates a nested loop geometry reminiscent of that described in analytical models of the heart \cite{kaplan1990fibrillation,richter1998phase}. This structure persists despite the plotted embedding corresponding to an ECG from a different patient than the one used to train the model. For the Old Faithful dataset, the model identifies a low-dimensional quasi-periodic attractor that is consistent with a long-speculated hypothesis that the geyser's nearly-regular dynamics arise from a strange attractor (spanning a small number of governing pressure and temperature state variables) \cite{nicholl1994old}. The dense regions of the attractor correspond to eruption events---which occur with consistent, stereotyped dynamics---while the diffuse, fan-like region of the attractor corresponds to the slow recovery period between firings, which has a broader range of dynamics and timings. For the electricity usage dataset, the embedding reveals a circular limit cycle consistent with a stable daily usage cycle, in agreement with other time-series analysis algorithms \cite{rangapuram2018deep}. For the mouse neuron spiking rate dataset, the model identifies a double-limit-cycle structure, consistent with higher-dimensional measurements suggesting that the neuronal dynamics lie on an intrinsic attractor manifold \cite{chaudhuri2019intrinsic}.

\section{Discussion}

We have introduced a method for reconstructing the attractor of a dynamical system, even when only low-dimensional time series measurements of the system are available. By comparing our technique to existing methods across a variety of complex time series, we have shown that our approach constructs informative, topologically-accurate embeddings that capture the intrinsic dimensionality of the original system. In practice, we are able to obtain strong results by tuning one hyperparameter, the regularizer strength $\lambda$, while adjusting the learning rate only to ensure convergence. Moreover, we empirically observe that the magnitude of $\lambda$ has a nonlinear effect on the structure of the embedding, providing a simple heuristic for tuning this hyperparameter in unsupervised settings. For this reason, we anticipate that our method can readily be applied to unknown time series datasets, which we have demonstrated using examples from areas spanning physiology, neuroscience, and geophysics. However, we note that our technique does introduce some tradeoffs compared to other methods of embedding time series: for simple or small datasets, lagged coordinates or eigen-time-delays are less computationally-expensive than our neural network based approach. However, for larger or more complex datasets, our method offers the same scaling advantages as other applications of neural networks to large datasets. Moreover, our general-purpose activity regularizer can be used with any network architecture and thus potentially other problem domains beyond time series embedding. Our technique and open-source implementation supports multidimensional time series, and in future work we hope to further draw upon classical results in the theory of chaotic systems in order to more directly relate quantitative properties of the learned attractors---such as the Lyapunov exponents and fractal dimension---to statistical features of the network's underlying representation of the time series. More broadly, we hope that our approach can inform efforts to learn differential equation models that describe latent dynamics \cite{schmidt2009distilling,champion2019data,cranmer2020discovering}, which have recently been shown to exhibit a tunable tradeoff between accuracy and parsimony \cite{udrescu2020ai}---an effect that may be mitigated by more constrained latent representations.

\clearpage
\newpage

\section*{Broader Impact}

While our work is motivated by conceptual and theoretical problems in the theory of nonlinear and chaotic dynamical systems, our method incurs comparable ethical and societal impacts to other techniques for mining time series data. While we focus primarily on datasets from the natural sciences, our technique can be used to identify recurring patterns and motifs within industrial datasets, such as subtle usage patterns of a utility, or fluctuations in consumer demand. Likewise, our technique is well-suited to the analysis of data from fitness trackers, in which a small number of measured dynamical variables (acceleration, elevation, heart rate, etc) are used as a proxy for a model of an individual's behavior (and even overall health status). Both of these examples represent cases in which the technique may identify latent factors about individuals that they might not anticipate being observable to a third-party---thus introducing concerns about privacy. Avoiding such problems first requires intervention at the level of the data provided to the technique (e.g., aggregating time series across individuals before training the model) or strict retention limitation policies (e.g., providing an overall health score, and then deleting the underlying model behind that score). For example, in our technique, the rows of the Hankel matrix used to train the model can be randomly sampled from a pool of user time series, resulting in an aggregated model that avoids modeling any particular user.

However, the ability of our method to identify latent dynamical variables also motivates potential positive societal impacts, particularly in regards to potential benefits for the analysis of physiological data. We show examples of non-trivial structure being extracted from cardiac measurements and neural activity; these structures may be used to better identify and detect anomalous dynamics, potentially improving health monitoring. More broadly, better identification of latent factors within time series data may allow for more principled identification and removal of features that undermine individual privacy, although this would require {\it ex post facto} analysis and interpretation of latent variables discovered using our technique.

\begin{ack}
We thank Chris Rycroft, Daniel Forger, Sigrid Keydana, Brian Matejek, Matthew Storm Bull, and Sharad Ramanathan for their comments on the manuscript. W. G. was supported by the NSF-Simons Center for Mathematical and Statistical Analysis of Biology at Harvard University, NSF Grant DMS 1764269, and the Harvard FAS Quantitative Biology Initiative. The author declares no competing interests.
\end{ack}

\bibliography{fnn_bib}
\bibliographystyle{naturemag}

\clearpage
\newpage

\tableofcontents
\renewcommand{\thetable}{S\arabic{table}}
\setcounter{table}{0}
\renewcommand{\thefigure}{S\arabic{figure}} 
\setcounter{figure}{0}
\renewcommand{\theequation}{A\arabic{equation}}
\setcounter{equation}{0}
\renewcommand{\thesubsection}{\Alph{subsection}}
\setcounter{subsection}{0}
\setcounter{section}{0}
\newpage

\section{Supplementary Code}

\noindent Code associated with this paper may be found at\\
 \url{https://github.com/williamgilpin/fnn}

\section{Calculation of the false-nearest-neighbor regularizer}

Our loss function $\mathcal{L}_{\text{FNN}}$ represents a variational formulation of the false-nearest-neighbors method, a popular heuristic for determining the appropriate embedding dimension $d_E$ when using the method of lags \cite{kennel1992determining}. The intuition behind the technique is that a $d$-dimensional embedding with too few dimensions will have many overlapping points, which will undergo large separation when the embedding is lifted to $d+1$. These points correspond to false neighbors, which only co-localize in $d$ dimensions due to having overlapping projections (Figure \ref{neighbors}). The traditional false-nearest-neighbors technique asserts that the true embedding dimension $d_E$ occurs when the fraction of false nearest neighbors first approaches zero as $d$ increases. 

Let $h \in \mathbb R^{B \times L}$ denote activations of a latent layer with $L$ units, generated when the network is given an input batch of size $B$. For the embedding problem studied here, $h$ corresponds to a partial embedding $\sim \hat Y$ generated from an input comprising $B$ length-$T$ rows randomly sampled from the full Hankel measurement matrix $X$. However, here we use general notation to emphasize that this regularizer can be applied to hidden layers in an arbitrary network.

We define the dimension-indexed, pairwise Euclidean distance $D \in \mathbb R^{B \times B \times L}$ among all points in the batch,
\[
D_{abm}^2 = \sum_{i=1}^m (h_{ai} - h_{bi})^2.
\]
This tensor describes the Euclidean distance between samples $a$ and $b$ when only the first $m$ latent dimensions are considered. Calculation of this quantity therefore breaks ordering invariance among the latent dimensions. 

We now define two related quantities: $\tilde D_{abm} \in \mathbb R^{B \times B \times L}$ corresponds to $D_{abm}$ sorted columnwise, while $\tilde D_{abm}' \in \mathbb R^{B \times B \times (L-1)}$ contains each column of $D_{abm}$ ordered by the sort order of the previous column. We calculate these quantities first by calculating the index tensor $g \in \mathbb{R}^{B \times B \times L}$, where each column $g_{a,:,m}$ contains the indices of all members of the batch sorted in ascending order of their relative distance from $a$ when only the first $m$ dimensions are considered. We then use $g$ to define
\[
\tilde D_{abm} = \sum_{\beta=1}^B \delta_{\beta, g_{abm}} D_{a\beta m}, \quad
\tilde D_{abm}' = \sum_{\beta=1}^B \delta_{\beta, g_{ab,m-1}} D_{a\beta m}.
\]
These quantities allow computation of the normalized change in distance to a given neighbor as $m$ increases, labelled by its relative distance, $S_{abm} = (\tilde{D}_{abm}'^2 - \tilde{D}_{abm}^2)/\tilde{D}_{abm}^2$, where $m \geq 2$.

A {\it false neighbor} is an $m-1$ dimensional near-neighbor that undergoes a jump greater than $R_{\text{tol}}$ when lifted to $m$ dimensions. We therefore define a binary tensor describing whether each point $a$ undergoes a jump of this magnitude in its $m^{th}$ dimension,
\[
R_{abm} =
\begin{cases} 
      1 & S_{abm} \geq R_{\text{tol}} \\
      0 & S_{abm} < R_{\text{tol}}
 \end{cases}.
\]
The threshold $R_{\text{tol}}$ can be chosen arbitrarily; in practice we find that it has little effect on our results, and so we set it to a constant value $R_{\text{tol}} = 10$ (a standard value) for all experiments \cite{kennel1992determining}. 

In regions of the attractor where the dynamics proceeds relatively quickly, the uniformly-spaced time series comprising $\hat Y$ undersamples the attractor. This can lead to points undergoing large shifts in position relative to the scale of the attractor as $m$ increases, leading to an additional criterion for whether a given point is considered a false neighbor. We define the characteristic size of the attractor in the first $m$ latent coordinates,
\[
\mathcal{R}_m^2 = \dfrac{1}{m\,B}\sum_{b=1}^B \sum_{i=1}^m (h_{bi} - \bar h_i)^2,
\]
where $\bar h_i = (1/B) \sum_{b=1}^B h_{bi}$. This quantity defines a second criterion,
\[
A_{akm} =
\begin{cases} 
      1 & \tilde D_{abm} \geq A_{\text{tol}} \mathcal{R}_m\\
      0 & \tilde D_{abm} < A_{\text{tol}} \mathcal{R}_m
 \end{cases}.
\]
The behavior of the regularizer does not strongly vary with $A_{\text{tol}}$, as long as this hyperparameter is set to a sufficiently large value. We therefore set $A_{\text{tol}} = 2.0$, a standard value in the literature, and keep it constant for all experiments.

We define the elementwise false neighbor matrix, which indicates points that satisfy either or both of these criteria,
\[
F_{abm} = \Theta(R_{abm} + A_{abm})
\]
where $\Theta$ denotes the left-continuous Heaviside step function, $\Theta(x) = 1, x>0$, $\Theta(x) = 0, x\leq0$. We next contract dimensionality by averaging this quantity $F_{abm}$ across both the batch and the set of $K$ nearest neighbors to $a$,
\[
\bar F_{m} = \dfrac{1}{K\,B} \sum_{k=1}^K \sum_{b=1}^B F_{kbm}.
\]
The hyperparameter $K$ determines how many neighbors are considered close enough to be informative about the topology of the attractor. Because varying this hyperparameter has a similar effect to changing $B$, we set $K = \max(1, \lceil 0.01 B \rceil)$ and otherwise leave this parameter constant; as with the original false-nearest-neighbors method, our approach performs well even when $K=1$ \cite{kennel1992determining}. Having obtained the dimension-wise fractional false neighbor count $\bar{F}_{m}$, we now calculate the false neighbor loss,
\[
\mathcal L_{\text{FNN}} = \sum_{m=2}^L (1- \bar{F}_{m}) \bar{h}_m^2.
\]
where $\bar{F}_{m},  \bar{h}_m$ and thus $\mathcal L_{\text{FNN}}$ implicitly depend on the batch activations $h$. Overall, $\mathcal L_{\text{FNN}}$ has the form of an activity regularizer acting on the latent coordinates. The overall loss function for the autoencoder is therefore
\[
\mathcal L(X, \hat X, \hat Y) = \| X - \hat X \|^2 + \lambda\, \mathcal{L}_{\text{FNN}}(\hat Y)
\]
where $\|.\|^2$ denotes the mean square error averaged across the batch, and $\lambda$ is a hyperparameter controlling the relative strength of the regularizer.

\begin{figure}
  \centering
  \includegraphics[width=\linewidth]{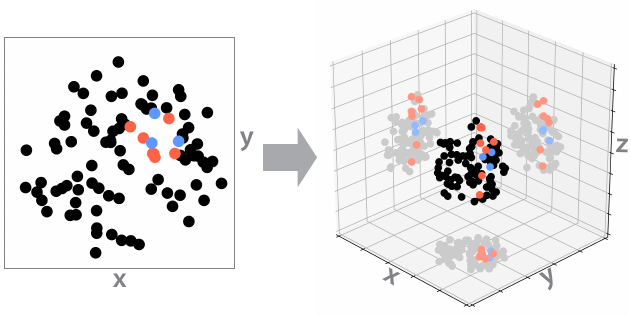}
  \caption{A set of near neighbors in a two-dimensional projection of three-dimensional point cloud (circled blue and red points). False neighbors (red) separate when the system is lifted to a higher dimension.}
  \label{neighbors}
\end{figure}

\section{Description of reference datasets}
\label{data_appendix}

{\bf Lorenz attractor.} The Lorenz equations are given by
\begin{eqnarray}
\dot x &=& \sigma(y - x) 	\\
\dot y &=& x(\rho - z) - y 	\\
\dot z &=& x y - \beta z
\end{eqnarray}
We use parameter values $\sigma=10$, $\rho=28$, $\beta=2.667$. The system is simulated for $500$ timesteps, with a stepsize $\Delta t = 0.004$. The system is then downsampled by a factor of $10$.  We fit the model using $x(t)$, which we divide into separate train, validation, and test datasets comprising $5000$ timepoints sampled from three trajectories with different initial conditions. To avoid transients, for each partition we select the last $5000$ timepoints from a $125000$ step trajectory. For stochastic simulations of this system, an uncorrelated white noise term $\xi(t)$, $\langle \xi(t) \xi(t') \rangle = \xi_0^2 \delta(t - t')$ is appended to each dynamical variable before integration, the integration timestep is decreased to $\Delta t = 0.0004$, and the integration output is downsampled by a factor of $100$.

{\bf R\"ossler attractor.} The R\"ossler attractor is given by
\begin{eqnarray}
\dot x &=&-y - z	\\ 
\dot y &=& x + ay\\
\dot z &=& b+z(x-c)
\end{eqnarray}
We use parameter values $a=0.2$, $b=0.2$, $c=5.7$, which produces a chaotic attractor with the shape of a M\"obius strip. The system is simulated for $2500$ timesteps, with a stepsize $\Delta t = 0.125$. The system is then downsampled by a factor of $10$.  We fit the model using $x(t)$, which we divide into separate train, validation, and test datasets corresponding to different initial conditions.

{\bf Ecological resource competition model.} We use a standard resource competition model, a variant of the Lotka-Volterra model that is commonly used to describe scenarios in which $n$ distinct species compete for a pool of $k$ distinct nutrients. We let $N_i(t)$ denote the abundance of species $i$, and $R_j(t)$ denote the availability of resource $j$. 
\begin{eqnarray}
\dot N_i &=& N_i \bigg(	\mu_i(R_1, ..., R_k) - m_i	\bigg)		\\
\dot R_j &=& D(S_j - R_j) - \sum_{i=1}^n c_{ji}\,\mu_i(R_1, ..., R_k) N_i	
\end{eqnarray}
where the species-specific growth rate is given by
\[
\mu_i(R_1, ..., R_k) = \min\bigg(	\frac{r_i R_1}{K_{1i} + R_1}, ..., \frac{r_i R_k}{K_{ki} + R_k}		\bigg).
\]

This model is strongly chaotic for a range of parameter values, and it was recently used to argue that chaotic dynamics may account for the surprising stability in long-term population abundances of competing phytoplankton species in the ocean \cite{huisman1999biodiversity}. We use parameter values from this study, which corresponds to $n=5$ species and $k=5$ resources. The full parameter values are: $D = 0.25$, $r_i = r = 1$, $m_i = m = 0.25$, $\v S = [6, 10, 14, 4, 9]$, 
$\v K = 
\begin{bmatrix} 
0.39 & 0.34 & 0.3 & 0.24 & 0.23\\
0.22& 0.39& 0.34& 0.3 & 0.27 \\
0.27& 0.22& 0.39& 0.34& 0.3  \\
0.3 & 0.24& 0.22& 0.39& 0.34 \\
0.34& 0.3 & 0.22& 0.2 & 0.39 
\end{bmatrix},
$

$\v c = 
\begin{bmatrix} 
0.04 &  0.04 &  0.07 &  0.04 &  0.04  \\ 
0.08 &  0.08 &  0.08 &  0.1  &  0.08  \\ 
0.1  &  0.1  &  0.1  &  0.1  &  0.14  \\ 
0.05 &  0.03 &  0.03 &  0.03 &  0.03  \\ 
0.07 &  0.09 &  0.07 &  0.07 &  0.07  
\end{bmatrix}.
$

We simulate this system for $200000$ units of time, with timestep $\Delta t = 0.1$. We discard the first $100000$ timepoints to eliminate any transients, and then downsample the time series by a factor of $10$. We fit the model using $R_1(t)$, which we divide into separate train, validation, and test datasets corresponding to different initial conditions.

{\bf Three-dimensional torus.} We parametrize a torus as a continuous-time, quasiperiodic dynamical system
\begin{eqnarray}
\dot x &=& -a n\sin(n t) cos(t) - (r + a\, \cos(n t)) \sin(t)	\\
\dot y &=& -a n \sin(n t)\sin(t) + (r + a\cos(n t))\cos(t)	\\
\dot z &=& a n \cos(n t)
\end{eqnarray}
where we use the parameters $r = 1$ (the outer radius), $a=1/2$ (the cross-sectional radius), $n=15.3$ (the winding number). Because $n$ is not an integer, trajectories of this system are non-recurring and quasiperiodic. The system is simulated for $2000$ timesteps, with a stepsize $\Delta t = 0.02$. The time series is then downsampled by a factor of $8$.  We fit the model using $x(t)$, which we divide into separate train, validation, and test datasets corresponding to different initial conditions.

{\bf Double pendulum experimental dataset.} We use an existing experimental dataset comprising a 400 fps video of a double pendulum experiment, recorded on a high-speed Phantom Miro EX2 camera \cite{asseman2018learning}. The video was segmented by the original authors, and the centroid positions were recorded for the pivot attachment to the wall, the joint between the first and second pendula, and the tip of the second pendulum. We convert this dataset into new time series corresponding to the angles that the first and second pendulum make with the vertical direction, $(\theta_1, \theta_2)$. These time series are then numerically differentiated, in order to produce a time series of the angular velocities $(\dot \theta_1, \dot \theta_2)$. For an ideal double pendulum, the four coordinates $(\dot \theta_1, \dot \theta_2, \theta_1, \theta_2)$ canonically parametrize the Hamiltonian of the system, and so these four coordinates are used as the definition of the attractor. However, we note that, for the experimental dataset, the time-averaged kinetic energy $T \propto \dot{\theta_1}^2 + \dot{\theta_2}^2$ gradually decreases throughout the course of the experiment. This additional coordinate was not included in the reference description of the attractor, due to its slow dynamics and non-stationarity, and so it constitutes an external, non-autonomous source of variation for which the model must compensate.

We downsample the raw time series by a factor of $3$ and us $\dot\theta_1(t)$ as the input to the model. For training and validation, we use the first and second sequences of $5000$ timepoints from the first experimental dataset. For testing, we use the first $5000$ timepoints from the second experimental dataset.

%

\section{Description of exploratory datasets}

{\bf Electrocardiogram.} We use recordings from the PhysioNet QT database, which comprises fifteen-minute, two-lead ECG recordings from multiple individuals \cite{laguna1997database,goldberger2000physiobank}. Measurements are spaced $0.004$ seconds apart. To remove high-frequency components, datasets were smoothed with a third-order Savitzky-Golay filter with a window size of $15$ timepoints. The datasets are then downsampled by a factor of $10$. For the analysis presented here, we use $10000$ datapoints (post-subsampling) from the dataset \texttt{sel102.dat} as training data, and for testing data we use $10000$ datapoints from the dataset \texttt{sel103.dat} (which corresponds to a different patient). 

{\bf Electricity usage.}  We use a dataset from the UCI machine learning database \cite{rangapuram2018deep,dua2019machine}, comprising residential power consumption by 321 Portuguese households between 2012 and 2014. Raw data is measured in units of kilowatts times the fifteen minute sampling increment. We create a consolidated dataset by taking the mean of all residences at each timepoint, adjusting the sample size as necessary at each timepoint to account for missing values for some households. We use the first, second, and last $10000$ timepoints training, validation, and testing data.

{\bf Geyser temperature measurements.} We use temperature recordings from the GeyserTimes database (\texttt{https://geysertimes.org/}), which consist of temperature readings from the main runoff pool of the Old Faithful geyser, located in Yellowstone National Park. Temperature measurements start on April 13, 2015 and occur in one-minute increments. The dataset was detrended by subtracting out a version of the data smoothed with a moving average over a one-day window, which effectively removes gradual effects like seasonal variation from the attractor. For the analysis presented in the main text, we use the first, second, and last $10000$ datapoints from the Old Faithful dataset as training, validation, and test datasets, respectively, corresponding to $\approx 400$ eruptions of the geyser.

{\bf Neural spiking.} We use a dataset from a recent study characterizing the intrinsic attractor manifold of neuron firings in freely-moving mice \cite{chaudhuri2019intrinsic}. The raw spike count data is available from the CRCNS database (\texttt{http://crcns.org/data-sets/thalamus/th-1}), and we process this data using the authors' included code and instructions, in order to generate time series corresponding to spiking rates for single neurons. We use the first, second, and last $10000$ timepoints training, validation, and testing data.

\section{Models}

We apply eigen-time-delay (ETD) embedding as in previous studies \cite{brunton2017chaos}, using principal component analysis as implemented in \texttt{scikit-learn} \cite{pedregosa2011scikit}. We apply time-structure independent component analysis (tICA) as implemented in the \texttt{MSMBuilder} software suite \cite{harrigan2017msmbuilder}. For numerical integration of chaotic systems, we use the LSODA method as implemented in \texttt{scipy} \cite{virtanen2019scipy}.

Autoencoders are implemented using TensorFlow \cite{abadi2016tensorflow}. The LSTM autoencoder has architecture: [Input-GN-LSTM(10)-BN]-[GN-LSTM(10)-BN-ELU-Output]. The multilayer perceptron has architecture: [Input-GN-FC(10)-BN-ELU-FC(10)-BN-ELU-FC(10)-BN]--[GN-FC(10)-BN-ELU-FC(10)-BN-ELU-FC(10)-BN-ELU-Output]. ELU denotes an exponential linear unit with default scale parameter $1.0$, BN denotes a BatchNorm layer, GN denotes a Gaussian noise regularization layer (active only during training) with default standard deviation $0.5$, and $FC$ denotes a fully-connected layer. $10$ hidden units are used in all cells, including for the latent space $L=10$, and network architecture or structural hyperparameters are kept the same across experiments. For both architectures, no activation is applied to the layer just before the latent layer, because the shape of the activation function is observed to constrain the range of values in latent space, consistent with prior studies \cite{otto2019linearly}.

\section{Extended description of similarity metrics}

{\bf Evaluation metrics.} We introduce several methods for comparing the original system $Y$ with its reconstruction $\hat Y$. We emphasize that this comparison does not occur during training (the autoencoder only sees one coordinate); rather, we use these metrics to assess how well our models can reconstruct known systems.

{\it 1. Dimension accuracy.} A basic, informative property of a dynamical system $\dot{\v{y}}(t)$ is its dimensionality, $d = \dim(\v{y})$, the minimum number of distinct variables necessary to fully specify the dynamics. Embeddings with $d_E < d$ discard essential information by collapsing independent coordinates, while embeddings with $d_E > d$ contain redundancy. We thus introduce a measure of embedding parsimony based on the effective number of latent coordinates present in the learned embedding. 
 
We equate the activity of a given latent dimension with its dimension-wise variance $\text{Var}(\hat{\v y})$, calculated across the ensemble of model inputs $\{\v{x}_i\}_1^N$. We compare the distribution of activity in the reconstruction $\hat Y$ to the original attractor $Y$, padding the dimensionality of the original attractor with zeros as needed:
\begin{equation}
\mathcal{S}_{dim} = 1 - \dfrac{||\text{SORT}(\text{Var}(\v y)) - \text{SORT}(\text{Var}(\hat{\v{y}}))||}{||\text{Var}(\v y) ||}.
\label{sdim}
\end{equation}
This quantity is maximized when the number of active latent dimensions, and their relative activity, matches that found in the original attractor. We further discuss this score, and general properties of the embedding dimension $d_E$, in the next section.

{\it 2. Procrustes distance.} Because a univariate measurement cannot contain information about the symmetry group or chirality of the full attractor, when computing pointwise similarity between the true and embedded attractors, we first align the two datasets using the Procrustes transform,
\[
P =\arg\min _{\tilde P }\|\tilde P \hat Y-Y\|_{F}\quad \text {s.t.} \quad \tilde{P}^{\top}\tilde{P} =I,
\]
where $I$ is the identity matrix. This transformation linearly registers the embedded attractor to the original attractor via translation, rotation, reflection, but {\it not} shear. For example, after this transformation, mirror images of a spiral would become congruent, whereas a sphere and ellipsoid would not. After calculating this transform, we compute the standard Euclidean distance, which we normalize to produce a similarity metric,
\[
\mathcal{S}_{\text{proc}} = 1 - \dfrac{|| P \hat Y - Y ||}{|| Y - \bar Y  ||}
\]
where the mean square error $||.||^2$ is averaged across the batch, and $\bar Y_k = \sum_{b=1}^N Y_{kb}$. This metric corresponds to a weighted variant of a classical attractor similarity measure \cite{diks1996detecting}. In addition to the mean-squared error, we also calculate the dynamic time warping (DTW) distance between $P \hat Y$ and $Y$, which yields similar results as $\mathcal S_{\text{proc}}$. 

{\it 3. Persistent Homology.} The persistence diagram for a point cloud measures the appearance or disappearance of essential topological features as a function of length scale. A length scale $\epsilon$ is fixed, and then all points are replaced by $\epsilon$-radius balls, the union of which defines a surface. Key topological features (e.g., holes, voids, and extrema) are then measured, the parameter $\epsilon$ is increased, and the process is repeated. This process produces a birth-death diagram for topological features parametrized by different length scales. We refer to a recent review \cite{edelsbrunner2008persistent} for further details of the technique. Here, we build upon recent results showing that the Wasserstein distance between two persistence diagrams can be used as a measure of topological similarity between two dynamical attractors \cite{venkataraman2016persistent,tran2019topological}. We express this quantity as a normalized similarity measure
\[
\mathcal{S}_{\text{homol}}(\mathcal{P}_Y, \mathcal{P}_{\hat{Y}}) = 1 - \dfrac{d_b(\mathcal{P}_Y, \mathcal{P}_{\hat{Y}})}{d_b(\mathcal{P}_Y, 0)}
\]
where $\mathcal{P}_Y, \mathcal{P}_{\hat{Y}}$ denote the persistence diagrams associated with the point clouds $Y$ and $\hat{Y}$, and the denominator denotes distance to a "null" diagram with no salient topological features. Two attractors will have a high Wasserstein similarity if they share essential topological features (such as holes, voids, and extrema). We compute birth-death persistence diagrams using the \texttt{Ripser} software package \cite{tralie2018ripser}, and we compute Wasserstein distances between diagrams using the \texttt{persim} software package \cite{scikittda2019}.

{\it 4. Local neighbor accuracy.} We seek to quantify whether points on $\hat Y$ are embedded in the same neighborhood as they are on $Y$, using simplex cross-mapping \cite{sugihara1990nonlinear,sugihara2012detecting}. We summarize this technique here: We pick a single datapoint $\hat{\v y_i}$ from the attractor $\hat Y$, and then find the set $\{ j \}_1^k$ comprising its $k$ nearest neighbors on $\hat Y$. Following standard practice, we use the minimum number of neighbors to form a bounding simplex, $k = d_E + 1$ \cite{sugihara1990nonlinear}. We then select the corresponding $\{ j \}_1^k$ points from the attractor $Y$, producing the set $\{\v y_j\}_1^k$. The centroid of $\{\v y_j\}_1^k$ is used to generate an estimate $\v{\tilde{y}_j}$ for the position of point $\v y_j$. The procedure is repeated for all values of $i$, and the difference between $\v{\tilde{y}_j}$ and $\v y_j$ averaged across all points is used as the distance measure between $\hat Y$ and $Y$. In order to generate a time-delayed prediction, a factor $\tau$ is added to the indices of all points in $\{ j \}_1^k$. Following previous work, we convert this distance into a similarity metric $\mathcal{S}_{\text{simp}}$ by normalizing by the dimensionwise-summed variance of the positions of all points in $Y$, and then subtracting the resulting quantity from one \cite{linderman2017bayesian}. Generally $\mathcal{S}_{\text{simp}}$ decreases smoothly with $\tau$, and so we report results for several values of $\tau$.

\begin{figure}[ht]
\vskip 0.2in
\begin{center}
\centerline{\includegraphics[width=\linewidth]{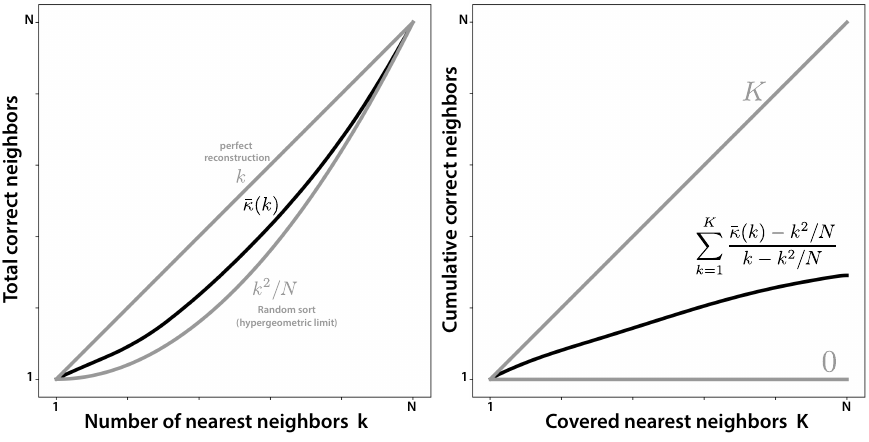}}
\caption{
Calculation of the nearest-neighbor coverage metric, $\mathcal{S}_{\text{nn}}$. (Left) the number of matching  $k$ nearest neighbors as a function of $k$ for two identical point clouds, an empirical reconstruction of a point cloud, and a cloud of random points (for which the fraction of matching nearest neighbors is given by the hypergeometric distribution). (Right) the cumulative sum of the quantities on the left, scaled to lie in the interval between the two values.
}
\label{coverage}
\end{center}
\vskip -0.2in
\end{figure}

{\it 5. Global neighbor coverage.} For the $i^{th}$ point of the $N$ embedded points in $\hat Y$, we define $\kappa_i(k)$ as the number $k$ nearest neighbors that correspond to true neighbors in the original dataset $Y$. For example, if the indices of the three closest neighbors to point $1$ in $Y$ are $11,14,29$ in order of relative distance, whereas its three closest neighbors are $11,29,15$ in $\hat Y$, then $\kappa_1(1) = 1, \kappa_1(2) = 1, \kappa_1(3)=2$. We average this quantity across all points in $\hat Y$, $\bar\kappa(k) = \sum_{b=1}^B \kappa_b(k)$. We note that, for a random shuffling of neighbors, $\kappa(k)$ is given by the hypergeometric distribution describing a random sample of $k$ objects from a collection of $N$ distinct objects without replacement, $\kappa(k) \sim f(N, N, k), \bar{\kappa}(k) = k^2/N$; in contrast, a set of perfectly matching neighbors will exhibit $\bar{\kappa}(k) = k$. We use these bounds to define the neighbor similarity as the area under the curve between the observed $\bar\kappa(k)$ and the random case, normalized by the best-case-scenario
\[
\mathcal{S}_{\text{nn}} = \dfrac{1}{N} \sum_{k=1}^{N - 1} \dfrac{\bar\kappa(k) - k^2/N}{k - k^2/N} 
\]
Similar to an ROC AUC, this metric depends on the fraction of correct neighbors within the closest $k$ neighbors, as the parameter $k$ is swept. We illustrate calculation of this quantity diagrammatically in Figure \ref{coverage}. 

{\it 6. Fractal dimension.} As an example of a physically-informative quantity that can be computed for an attractor, but not a raw time series, we compare the correlation dimension (a type of fractal dimension) of the original attractor $c_Y$ and its reconstruction $c_{\hat Y}$ using the symmetric mean absolute percent error
\[
\mathcal S_{\text{corr}}(c_Y, c_{\hat Y}) = 1 - \dfrac{\abs{c_Y -  c_{\hat Y}}}{\abs{c_Y} -  \abs{c_{\hat Y}}}.
\]
We use the correlation dimension instead of related physical properties (such as the Lyapunov exponent, or Kolmogorov-Sinai entropy) because, unlike other properties, the correlation dimension can be robustly measured in a parameter-free manner, without random subsampling of points \cite{grassberger1983measuring}.

\section{Additional Experiments}

\subsection{Application to time series clustering}

Outside of physics, a significant application of attractor reconstruction lies in improving the representation and featurization of time series datasets \cite{karl2016deep,durbin2012time,rangapuram2018deep}. We apply our technique to four time series classification tasks from different application domains: (1) a synthetic dataset consisting of the $x$ coordinate of simulations of the Lorenz equations with different initial conditions, labelled by the exact values of the parameters defining the equations, (2) the first principle component of the body shape of crawling {\it C. elegans} worms, labelled by the genetic mutant; (3) electrocardiogram recordings of patients undergoing either standing, walking, or jumping; (4) electroencephalogram measurements of patients imagining one of two possible movements \cite{yemini2013database,goldberger2000physiobank,lal2005methods}. We do not tune hyperparameters, and instead use the same default hyperparameters used to train the Lorenz attractor in the previous experiments. We use a $1$-nearest-neighbor classifier with dynamic time warping (a standard baseline for time series classification) \cite{bagnall2017great}, and summarize our results in Table \ref{class}. Across a variety of data sources and numbers of classes, classifiers using attractors obtained from our method achieve higher balanced accuracy than classifiers trained on the bare time series, or that use alternative embedding techniques. We obtain these results with no hyperparameter tuning, demonstrating that our method can generically extract meaningful features at each point in a time series---suggesting potential application of our approach as an initial featurization stage for general time series analysis techniques. 

\begin{table}[ht]
\caption{The balanced classification accuracy for different time series. The number of classes in each dataset is indicated in parentheses.}
\label{class}
\vskip 0.15in
\begin{center}
\begin{small}
\begin{sc}
\begin{tabular}{lccccr}
\toprule
Dataset & Raw & tICA &ETD  & LSTM  & LSTM-fnn \\
\midrule
\midrule
Lorenz ($8$)&0.18 & 0.22 & 0.18 &0.21 & {\bf 0.23}\\ 
Worm ($5$) & 0.52&0.45 & 0.39 &0.60 & {\bf 0.61}	\\ 
ECG  ($3$) &0.40& 0.20 & {\bf 0.47} &0.40& {\bf 0.47}	\\ 
EEG($3$) &0.46 &0.43 & 0.43 &0.44 & {\bf0.51}	\\ 
\bottomrule
\end{tabular}
\end{sc}
\end{small}
\end{center}
\vskip -0.1in
\end{table}

\subsection{Consistency and repeatability}

\begin{figure}[ht]
\vskip 0.2in
\begin{center}
\centerline{\includegraphics[width=\linewidth]{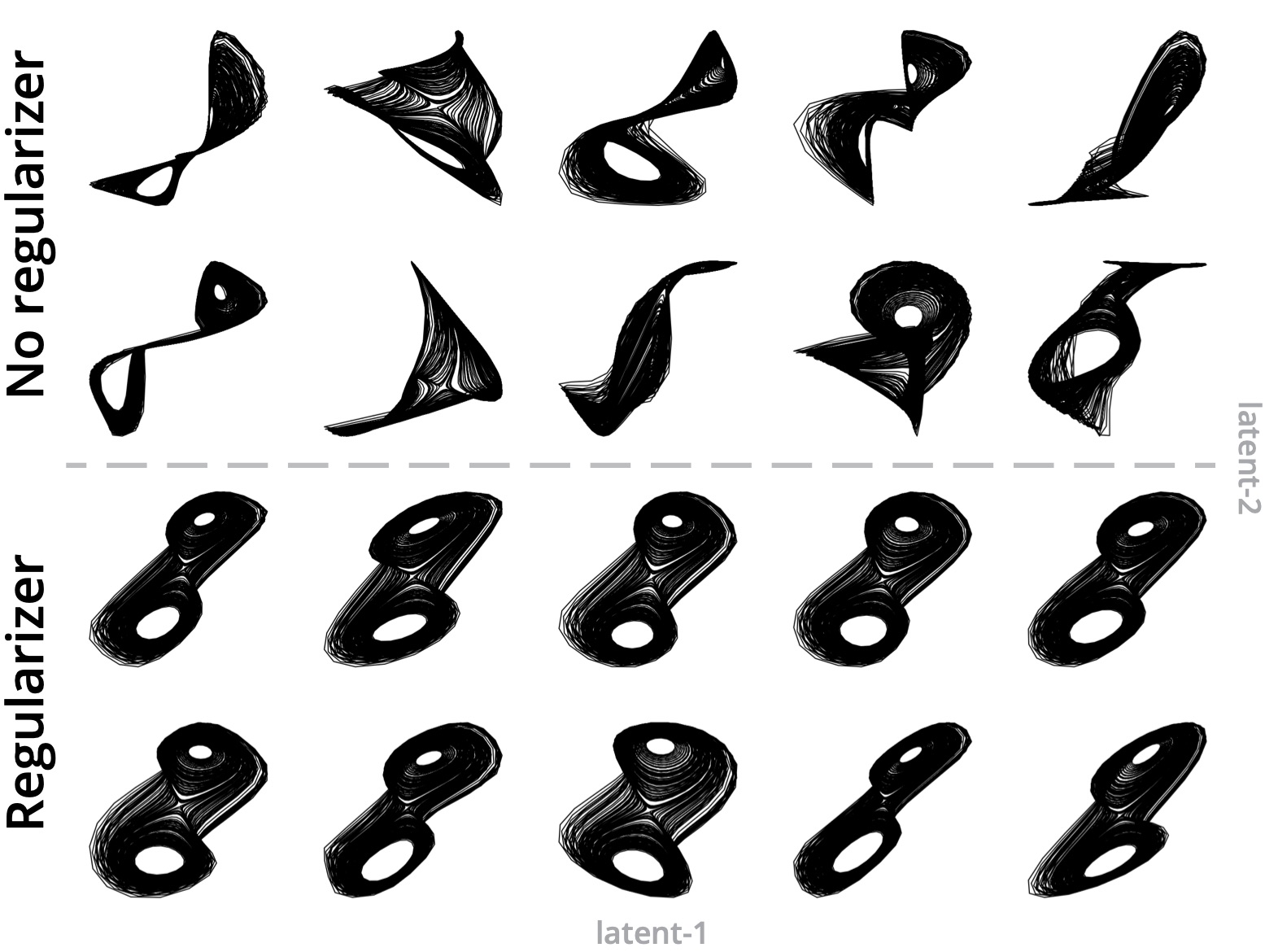}}
\caption{
An ensemble of reconstructed attractors for the Lorenz dataset, generated by models with different initial random weight initializations but identical hyperparameters. Upper portion of the plot shows models with no regularizer, and lower portion shows models with the false-nearest-neighbors regularizer. Before plotting, attractors were aligned using the Procrustes transform in order to remove random rotations.
}
\label{consist}
\end{center}
\vskip -0.2in
\end{figure}

We evaluate the repeatability and consistency of the learned representations by training an ensemble of models on the Lorenz dataset. All hyperparameters are held constant, and the only difference across replicates is the random weight initialization. As a baseline, we also trained a set of models with no false-neighbors regularization. Example embeddings of the test data for models with and without regularization are shown in Figure \ref{consist}. Before plotting, the Procrustes transform was used to remove random rotations. 

The figure demonstrates the regularizer produces significantly more consistent embeddings across replications, implying that the regularizer successfully constrains the space of latent representations. We quantify this effect by computing the pairwise topological similarity $\mathcal{S}_{\text{homol}}$ among all replicates (Table \ref{consist_table}), and we observe that the median topological similarity is larger for the regularized models.

\begin{table}[ht]
\caption{The median and standard error of the median across $20$ replicate models.}
\label{consist_table}
\vskip 0.15in
\begin{center}
\begin{small}
\begin{sc}
\begin{tabular}{lcc}
\toprule
\null & LSTM  & LSTM-fnn \\
\midrule
\midrule
$\langle \mathcal{S}_{\text{homol}}\rangle$ & $ 0.09 \pm 0.05$  &$0.21 \pm 0.07$ \\
\bottomrule
\end{tabular}
\end{sc}
\end{small}
\end{center}
\vskip -0.1in
\end{table}

\subsection{Effect of regularizer on alternate models}

We repeat the experiment (described in the main text) in which the regularizer strength $\lambda$ is varied, and show similar results for both the LSTM and the MLP autoencoders in Figure \ref{char}. 

\begin{figure}[ht]
\vskip 0.2in
\begin{center}
\centerline{\includegraphics[width=\linewidth]{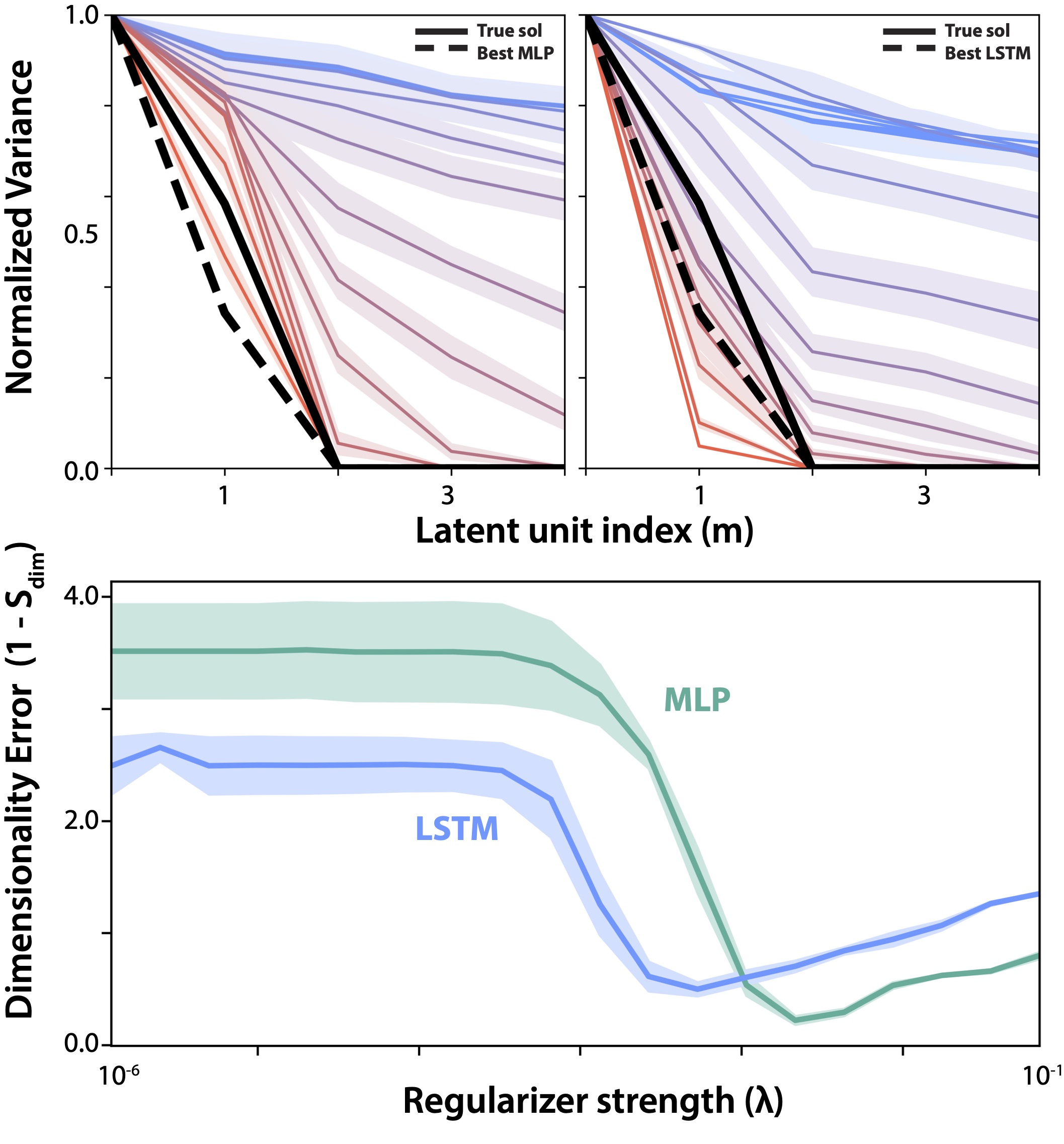}}
\caption{
(A) The distribution of normalized latent variances as a function of regularizer strength from $\lambda = 0$ (blue) to $\lambda = 0.1$ (red), with the normalized variance for the full solution (solid black line) and for the final best-performing model (dashed black line). (B) The dimensionality error $1 - \mathcal{S}_{\text{dim}}$ as a function of $\lambda$. Error ranges correspond to $5$ replicates.
}
\label{char}
\end{center}
\vskip -0.2in
\end{figure}

\subsection{Dimension error versus regularizer strength for pendulum dataset} 

\begin{figure}[ht]
\vskip 0.2in
\begin{center}
\centerline{\includegraphics[width=\linewidth]{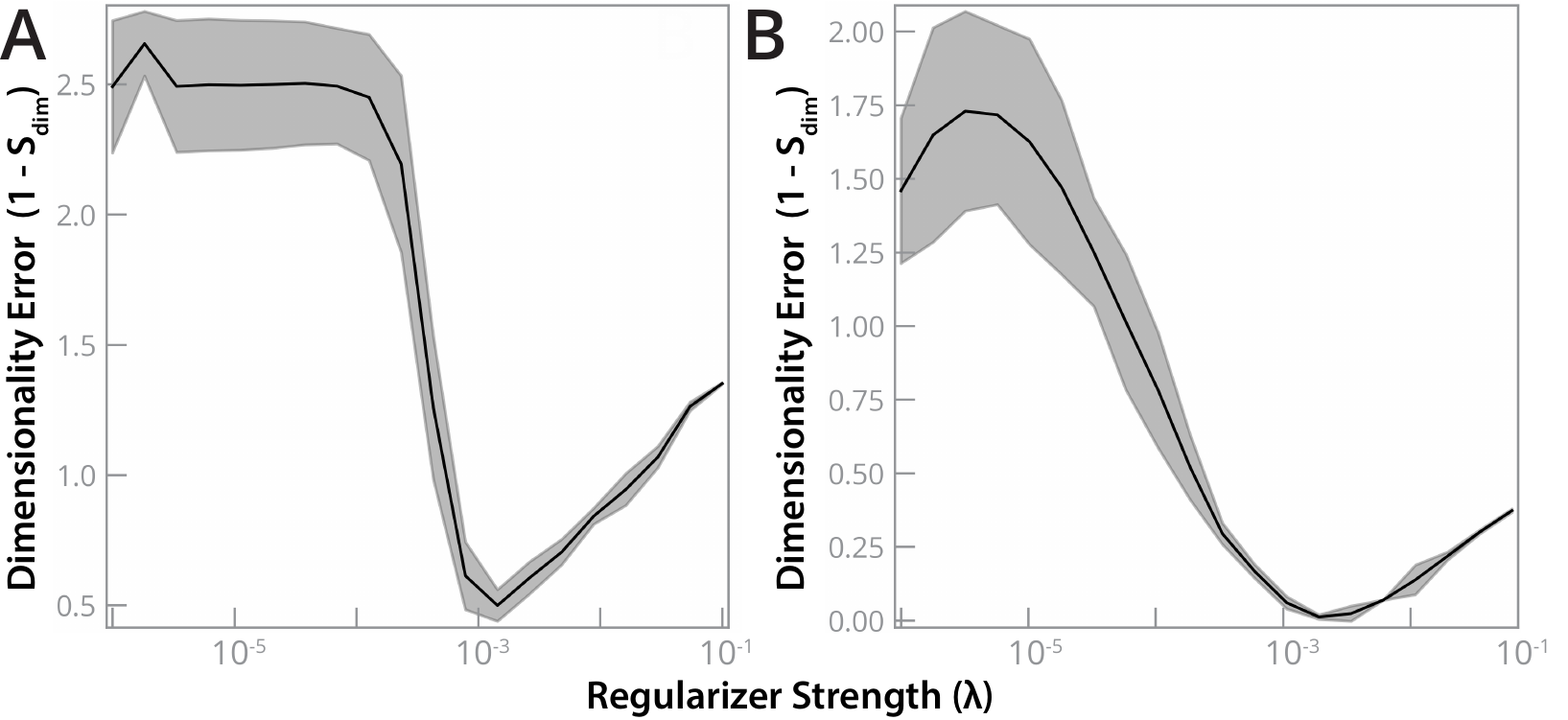}}
\caption{The final dimensionality error versus regularizer strength for the Lorenz dataset (A) and the double pendulum dataset (B). Error ranges correspond to $5$ replicates.
}
\label{pend_reg}
\end{center}
\vskip -0.2in
\end{figure}

In Figure \ref{pend_reg}, we repeat the experiment (described in the main text) in which we vary the regularizer strength ($\lambda$), and we observe that the final dimensionality error $\mathcal{S}_{\text{dim}}$ exhibits a similar nonlinear dependence on $\lambda$ for both the double pendulum and Lorenz datasets.

\subsection{Comparison to a vanilla activity regularizer} 

We also compare the false-neighbor regularizer to a standard L1 activity regularizer (across a variety of different regularizer strengths), and find that the false-neighbor regularizer shows improvement across the different metrics used in the main text.

\begin{figure}[ht]
\vskip 0.2in
\begin{center}
\centerline{\includegraphics[width=\linewidth]{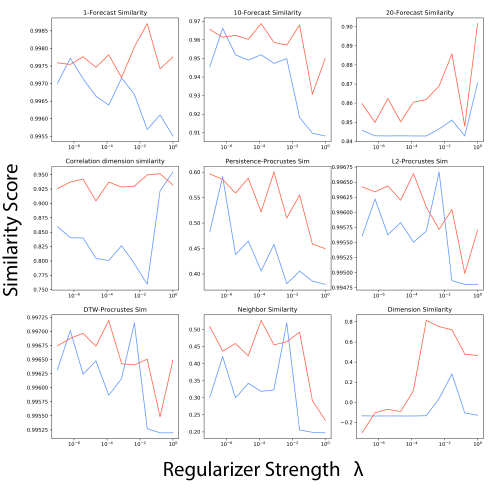}}
\caption{
Reconstruction accuracies for an LSTM model on the Lorenz dataset, using the false-nearest-neighbors regularizer (red) and a standard L1 activity regularizer (blue) on the latent units. While the regularizer strength is varied, all other hyperparameters are held constant at the values used in other experiments.
}
\label{char}
\end{center}
\vskip -0.2in
\end{figure}

\subsection{Dependence on the number of latent units} 

\begin{figure}[ht]
\vskip 0.2in
\begin{center}
\centerline{\includegraphics[width=\linewidth]{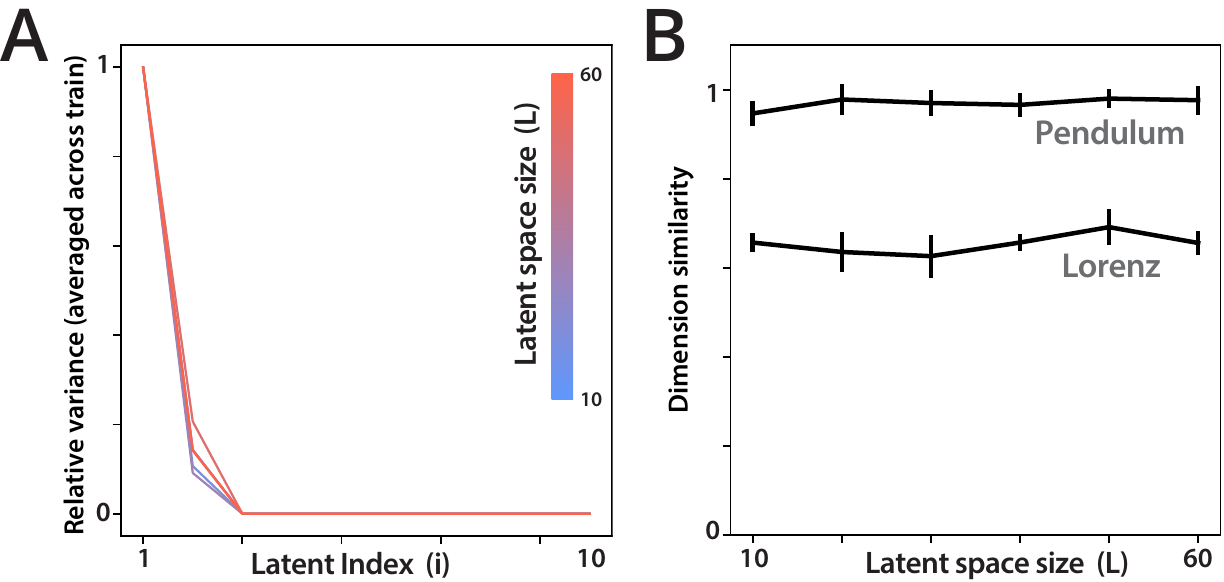}}
\caption{
(A) Activity patterns of the first $10$ latent units for autoencoders trained with $L=10,20,30,40,50$ units in the latent space (blue to red). (B) The dimension accuracy score vs $L$
}
\label{latent}
\end{center}
\vskip -0.2in
\end{figure}

In order to determine whether the hyperparameter $L$ influences the learned representations, we trained replicate autoencoders with the false-nearest-neighbor regularizer for $L=10,20,30,40,50$ latent units (Figure \ref{latent}). For all experiments, we used the Lorenz dataset with the same hyperparameters as were used in the main text. The figure shows that the relative activity of each latent unit after training is independent of $L$, implying that the network successfully learns to ignore excess latent dimensionality. Therefore, we argue that autoencoders trained with the false-nearest-neighbor regularizer will learn consistent representations that are determined primarily by the dataset, and by the relative strength of the regularizer $\lambda$.

\subsection{Forecasting comparison to previous autoencoders} 

\begin{figure}[ht]
\vskip 0.2in
\begin{center}
\centerline{\includegraphics[width=\linewidth]{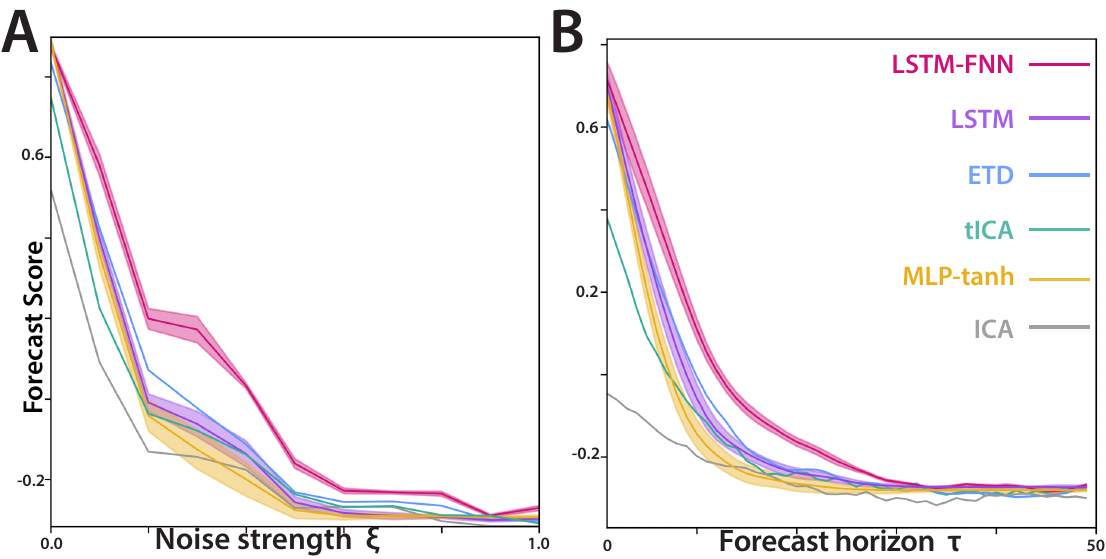}}
\caption{
(A) The cross-mapping forecast accuracy as a function of noise strength $\xi_0$ (with constant $\tau = 20$). (B) The cross-mapping forecast accuracy versus forecasting horizon $\tau$ (with constant $\xi_0 = 0.5$). Ranges correspond to standard error across $5$ random initializations.
}
\label{baseline}
\end{center}
\vskip -0.2in
\end{figure}

Previous work has proposed using a one-layer autoencoder with $\tanh$ activation in order to embed strange attractors, with the reconstruction loss serving as the sole loss function \cite{jiang2017state}. This prior work primarily focused on a forecasting task similar to our "noisy forecasting" experiments---the main difference being that we study simulations of stochastic differential equations (i.e. deterministic systems with added stochastic forcing), while the earlier work focuses on noisy measurements of a deterministic system. We re-implemented this earlier architecture and loss function, and compared it to our experiments with the stochastic Lorenz dataset (Figure \ref{baseline}, yellow traces). We find that the prior model performs slightly worse than our baseline unregularized LSTM, further underscoring the importance of the false-nearest-neighbors regularizer in generating consistent, predictive representations.

\section{All attractor comparison results}

Table \ref{all} shows the full results of attractor comparison experiments with all datasets, models, and metrics.

\begin{table*}[ht]
\caption{Results for five datasets with known attractors. Errors correspond to standard errors over $5$ replicates with random initial weights.}
\label{all}
\vskip 0.15in
\begin{center}
\begin{small}
\begin{sc}
\begin{tabular}{lccccccr}
\toprule
Metric & ICA & tICA & ETD & MLP & LSTM & MLP-fnn & LSTM-fnn \\
\midrule
\midrule
Lorenz \\
\midrule
$\mathcal{S}_{\text{simp}}$	&0.42	&0.74	&0.82	&0.79\,$\pm$\,0.03	&0.82\,$\pm$\,0.02	&0.81\,$\pm$\,0.03	&	0.93$\pm$\,0.02	\\
$\mathcal{S}_{\text{corr}}$	&0.992	&0.985	& 0.978	&0.91\,$\pm$\,0.01	&0.87\,$\pm$\,0.02	&0.953\,$\pm$\,0.009& 0.98\,$\pm$\,0.02	\\
$\mathcal{S}_{\text{homol}}$	&0.049	&0.123	& 0.668	&0.01\,$\pm$\,0.01	&0.04\,$\pm$\,0.03	& 0.47\,$\pm$\,0.05	&0.3\,$\pm$\,0.1	\\
$\mathcal{S}_{\text{proc}}$	&-0.015	&0.037	&0.212	&0.09\,$\pm$\,0.05	&0.20\,$\pm$\,0.03	&0.23\,$\pm$\,0.08	& 0.37\,$\pm$\,0.02	\\
$\mathcal{S}_{\text{dtw}}$	&0.237	&0.21	&0.27	&0.25\,$\pm$\,0.04	&0.31\,$\pm$\,0.03	&0.39\,$\pm$\,0.09	& 0.47\,$\pm$\,0.03	\\
$\mathcal{S}_{\text{nn}}$&0.277	&0.296	&0.384	&0.25\,$\pm$\,0.06	&0.25\,$\pm$\,0.06	&0.40\,$\pm$\,0.02	& 0.40\,$\pm$\,0.02	\\
$\mathcal{S}_{\text{dim}}$ 	&0.171	&0.394	& 0.628	&-0.4\,$\pm$\,0.1	&-0.06\,$\pm$\,0.08	& 0.88\,$\pm$\,0.02	&0.66\,$\pm$\,0.01	\\
\midrule
Double Pendulum \\
\midrule
$\mathcal{S}_{\text{simp}}$	&-0.24	&-0.06	& 0.30	&0.35\,$\pm$\,0.03	&0.36\,$\pm$\,0.02	&{ 0.39\,$\pm$\,0.02}	&{ 0.41\,$\pm$\,0.01}	\\
$\mathcal{S}_{\text{corr}}$	&0.985	&0.861	&0.822	&0.96\,$\pm$\,0.01	&{ 0.966\,$\pm$\,0.006}	&0.951\,$\pm$\,0.003	&{ 0.986\,$\pm$\,0.008}	\\
$\mathcal{S}_{\text{homol}}$	&0.191	&0.202	&0.176	&0.18\,$\pm$\,0.03	&0.19\,$\pm$\,0.02	&{ 0.26\,$\pm$\,0.03}	&{ 0.25\,$\pm$\,0.02}	\\
$\mathcal{S}_{\text{proc}}$	&0.002	&0.001	&0.003	&0.013\,$\pm$\,0.004	&0.008\,$\pm$\,0.005	&{ 0.013\,$\pm$\,0.003}	&{ 0.016\,$\pm$\,0.006}	\\
$\mathcal{S}_{\text{dtw}}$	&0.026	&0.069	&{ 0.108}	&0.11\,$\pm$\,0.02	&0.11\,$\pm$\,0.01	&{ 0.136\,$\pm$\,0.009}	&{ 0.132\,$\pm$\,0.008}	\\
$\mathcal{S}_{\text{nn}}$&0.019	&0.031	&0.055	&0.041\,$\pm$\,0.003	&0.042\,$\pm$\,0.002	&0.05\,$\pm$\,0.001	&{ 0.060\,$\pm$\,0.001}	\\
$\mathcal{S}_{\text{dim}}$ 	&-1.772	&-1.914	&0.927	&-0.6\,$\pm$\,0.2	&-0.8\,$\pm$\,0.3	&0.801\,$\pm$\,0.006	&{ 0.97\,$\pm$\,0.01}	\\
\midrule
Ecosystem \\
\midrule
$\mathcal{S}_{\text{simp}}$	&0.527	&0.532	&0.525	&0.89\,$\pm$\,0.02	&{ 0.92\,$\pm$\,0.03}	&{ 0.95\,$\pm$\,0.02}	&{ 0.93\,$\pm$\,0.03}	\\
$\mathcal{S}_{\text{corr}}$	&0.856	&{ 0.890}	& 0.820	&0.876\,$\pm$\,0.004	&0.877\,$\pm$\,0.004	&{ 0.904\,$\pm$\,0.009}	&{ 0.888\,$\pm$\,0.003}	\\
$\mathcal{S}_{\text{homol}}$	&0.185	&0.066	&0.256	&0.17\,$\pm$\,0.03	&0.09\,$\pm$\,0.04	&{ 0.36\,$\pm$\,0.03}	&{ 0.33\,$\pm$\,0.03}	\\
$\mathcal{S}_{\text{proc}}$	&{ 0.055}	&0.024	&0.025	&-0.01\,$\pm$\,0.03	&-0.1\,$\pm$\,0.05	&{ 0.04\,$\pm$\,0.03}	&{ 0.08\,$\pm$\,0.05}	\\
$\mathcal{S}_{\text{dtw}}$	&0.111	&0.115	&0.051	&{ 0.11\,$\pm$\,0.02}	&0.05\,$\pm$\,0.03	&{ 0.12\,$\pm$\,0.02}	&{ 0.15\,$\pm$\,0.03}	\\
$\mathcal{S}_{\text{nn}}$&0.133	&0.133	&0.146	&0.304\,$\pm$\,0.005	&0.304\,$\pm$\,0.004	&0.30\,$\pm$\,0.03	&{ 0.313\,$\pm$\,0.005}	\\
$\mathcal{S}_{\text{dim}}$ 	&-0.882	&0.60	&0.664	&0.38\,$\pm$\,0.08	&0.51\,$\pm$\,0.05	&{ 0.90\,$\pm$\,0.02}	&{0.92\,$\pm$\,0.02}	\\
\midrule
Torus \\
\midrule
$\mathcal{S}_{\text{simp}}$	&0.984	&0.996	&0.994	&{ 0.998\,$\pm$\,0.001}	&{ 0.999\,$\pm$\,0.001}	&{ 0.999\,$\pm$\,0.001}	&{ 0.998\,$\pm$\,0.002}	\\
$\mathcal{S}_{\text{corr}}$	&{ 0.994}	&0.952	&{ 0.993}	&0.982\,$\pm$\,0.006	&0.87\,$\pm$\,0.03	&{ 0.994\,$\pm$\,0.004}	&{ 0.99\,$\pm$\,0.01}	\\
$\mathcal{S}_{\text{homol}}$	&0.001	&-1.442	&-0.827	& -0.6\,$\pm$\,0.06	&-0.4\,$\pm$\,0.2	&-0.3\,$\pm$\,0.2	&{ 0.33\,$\pm$\,0.09}	\\
$\mathcal{S}_{\text{proc}}$	&0.157	&-0.102	&-0.008	&0.1\,$\pm$\,0.1	&-0.07\,$\pm$\,0.08	&{ 0.4\,$\pm$\,0.1}	&{ 0.4\,$\pm$\,0.1}	\\
$\mathcal{S}_{\text{dtw}}$	&0.403	&0.292	&{ 0.586}	&0.24\,$\pm$\,0.07	&0.19\,$\pm$\,0.07	&{ 0.60\,$\pm$\,0.08}	&{ 0.50\,$\pm$\,0.09}	\\
$\mathcal{S}_{\text{nn}}$&0.269	&0.194	&{ 0.444}	&0.28\,$\pm$\,0.03	&0.28\,$\pm$\,0.01	&{ 0.42\,$\pm$\,0.01}	&{ 0.45\,$\pm$\,0.02}	\\
$\mathcal{S}_{\text{dim}}$ 	&-0.619	&-0.652	&0.722	&0.1\,$\pm$\,0.1	&-0.3\,$\pm$\,0.3	&{ 0.96\,$\pm$\,0.04}	&0.71\,$\pm$\,0.01	\\
\midrule
R\"ossler \\
\midrule
$\mathcal{S}_{\text{simp}}$	&0.988	&0.997	& 0.997	& { 0.999\,$\pm$\,0.001}& 0.997\,$\pm$\,0.001	&{ 0.999\,$\pm$\,0.001}	&{ 0.999\,$\pm$\,0.001}	\\
$\mathcal{S}_{\text{corr}}$	&0.771	&0.994	&{ 0.999}	&0.94\,$\pm$\,0.02	&0.87\,$\pm$\,0.03	&0.985\,$\pm$\,0.003	&{ 0.997\,$\pm$\,0.003}	\\
$\mathcal{S}_{\text{homol}}$	&0.001	&0.06	&{ 0.501}	&0.08\,$\pm$\,0.04	&0.08\,$\pm$\,0.07	&0.27\,$\pm$\,0.04	&0.55\,$\pm$\,0.07\\
$\mathcal{S}_{\text{proc}}$	&0.123	&-0.002	&0.027	&0.01\,$\pm$\,0.09	&{ 0.33\,$\pm$\,0.04}	&{ 0.3\,$\pm$\,0.1}	&{ 0.25\,$\pm$\,0.06}	\\
$\mathcal{S}_{\text{dtw}}$	&0.351	&0.547	&0.527	&0.23\,$\pm$\,0.07	&0.43\,$\pm$\,0.05	&{ 0.52\,$\pm$\,0.09}	&{ 0.62\,$\pm$\,0.05}	\\
$\mathcal{S}_{\text{nn}}$&0.332	&{ 0.742}	&{ 0.762}	&0.43\,$\pm$\,0.03	&0.42\,$\pm$\,0.03	&0.64\,$\pm$\,0.01	&{ 0.75\,$\pm$\,0.06}	\\
$\mathcal{S}_{\text{dim}}$ 	&-0.48	&0.423	&{ 0.727}	&0.64\,$\pm$\,0.04	&0.5\,$\pm$\,0.1	&{ 0.694\,$\pm$\,0.05}	 &{ 0.753\,$\pm$\,0.08}	\\
\bottomrule
\end{tabular}
\end{sc}
\end{small}
\end{center}
\vskip -0.1in
\end{table*}

\clearpage

\bibliography{fnn_bib}

%
%
%
%

\end{document}